# Learning When Training Data are Costly: The Effect of Class Distribution on Tree Induction

**Gary M. Weiss**                                                    GMWEISS@ATT.COM
*AT&T Labs, 30 Knightsbridge Road*
*Piscataway, NJ 08854  USA*

**Foster Provost**                                                  FPROVOST@STERN.NYU.EDU
*New York University, Stern School of Business*
*44 W. 4th St., New York, NY 10012  USA*

## Abstract

For large, real-world inductive learning problems, the number of training examples often must be limited due to the costs associated with procuring, preparing, and storing the training examples and/or the computational costs associated with learning from them. In such circumstances, one question of practical importance is: if only *n* training examples can be selected, in what proportion should the classes be represented? In this article we help to answer this question by analyzing, for a fixed training-set size, the relationship between the class distribution of the training data and the performance of classification trees induced from these data. We study twenty-six data sets and, for each, determine the best class distribution for learning. The naturally occurring class distribution is shown to generally perform well when classifier performance is evaluated using undifferentiated error rate (0/1 loss). However, when the area under the ROC curve is used to evaluate classifier performance, a balanced distribution is shown to perform well. Since neither of these choices for class distribution always generates the best-performing classifier, we introduce a "budget-sensitive" progressive sampling algorithm for selecting training examples based on the class associated with each example. An empirical analysis of this algorithm shows that the class distribution of the resulting training set yields classifiers with good (nearly-optimal) classification performance.

## 1. Introduction

In many real-world situations the number of training examples must be limited because obtaining examples in a form suitable for learning may be costly and/or learning from these examples may be costly. These costs include the cost of obtaining the raw data, cleaning the data, storing the data, and transforming the data into a representation suitable for learning, as well as the cost of computer hardware, the cost associated with the time it takes to learn from the data, and the "opportunity cost" associated with suboptimal learning from extremely large data sets due to limited computational resources (Turney, 2000). When these costs make it necessary to limit the amount of training data, an important question is: in what proportion should the classes be represented in the training data? In answering this question, this article makes two main contributions. It addresses (for classification-tree induction) the practical problem of how to select the class distribution of the training data when the amount of training data must be limited, and, by providing a detailed empirical study of the effect of class distribution on classifier performance, it provides a better understanding of the role of class distribution in learning.

Some practitioners believe that the naturally occurring marginal class distribution should be used for learning, so that new examples will be classified using a model built from the same underlying distribution. Other practitioners believe that the training set should contain an increased percentage of minority-class examples, because otherwise the induced classifier will not classify minority-class examples well. This latter viewpoint is expressed by the statement, "if the sample size is fixed, a balanced sample will usually produce more accurate predictions than an unbalanced 5%/95% split" (SAS, 2001). However, we are aware of no thorough prior empirical study of the relationship between the class distribution of the training data and classifier performance, so neither of these views has been validated and the choice of class distribution often is made arbitrarily—and with little understanding of the consequences. In this article we provide a thorough study of the relationship between class distribution and classifier performance and provide guidelines—as well as a progressive sampling algorithm—for determining a "good" class distribution to use for learning.

There are two situations where the research described in this article is of direct practical use. When the training data must be limited due to the cost of learning from the data, then our results—and the guidelines we establish—can help to determine the class distribution that should be used for the training data. In this case, these guidelines determine how many examples of each class to *omit* from the training set so that the cost of learning is acceptable. The second scenario is when training examples are costly to procure so that the number of training examples must be limited. In this case the research presented in this article can be used to determine the proportion of training examples belonging to each class that should be *procured* in order to maximize classifier performance. Note that this assumes that one can select examples belonging to a specific class. This situation occurs in a variety of situations, such as when the examples belonging to each class are produced or stored separately or when the main cost is due to transforming the raw data into a form suitable for learning rather than the cost of obtaining the raw, labeled, data.

Fraud detection (Fawcett & Provost, 1997) provides one example where training instances belonging to each class come from different sources and may be procured independently by class. Typically, after a bill has been paid, any transactions credited as being fraudulent are stored separately from legitimate transactions. Furthermore transactions credited to a customer as being fraudulent may in fact have been legitimate, and so these transactions must undergo a verification process before being used as training data.

In other situations, labeled raw data can be obtained very cheaply, but it is the process of forming *usable* training examples from the raw data that is expensive. As an example, consider the *phone* data set, one of the twenty-six data sets analyzed in this article. This data set is used to learn to classify whether a phone line is associated with a business or a residential customer. The data set is constructed from low-level call-detail records that describe a phone call, where each record includes the originating and terminating phone numbers, the time the call was made, and the day of week and duration of the call. There may be hundreds or even thousands of call-detail records associated with a given phone line, all of which must be summarized into a single training example. Billions of call-detail records, covering hundreds of millions of phone lines, potentially are available for learning. Because of the effort associated with loading data from dozens of computer tapes, disk-space limitations and the enormous processing time required to summarize the raw data, it is not feasible to construct a data set using all available raw data. Consequently, the number of usable training examples must be limited. In this case this was done based on the class associated with each phone line—which is known. The phone data set was limited to include approximately 650,000 training examples, which were generated from approximately 600 million call-detail records. Because huge transaction-oriented databases are now routinely being used for learning, we expect that the number of training examples will also need to be limited in many of these cases.

The remainder of this article is organized as follows. Section 2 introduces terminology that will be used throughout this article. Section 3 describes how to adjust a classifier to compensate for changes made to the class distribution of the training set, so that the generated classifier is not improperly biased. The experimental methodology and the twenty-six benchmark data sets analyzed in this article are described in Section 4. In Section 5 the performance of the classifiers induced from the twenty-six naturally unbalanced data sets is analyzed, in order to show how class distribution affects the behavior and performance of the induced classifiers. Section 6, which includes our main empirical results, analyzes how varying the class distribution of the training data affects classifier performance. Section 7 then describes a progressive sampling algorithm for selecting training examples, such that the resulting class distribution yields classifiers that perform well. Related research is described in Section 8 and limitations of our research and future research directions are discussed in Section 9. The main lessons learned from our research are summarized in Section 10.

## 2. Background and Notation

Let $x$ be an instance drawn from some fixed distribution $D$. Every instance $x$ is mapped (perhaps probabilistically) to a class $C \in \{p, n\}$ by the function $c$, where $c$ represents the true, but unknown, classification function.[1] Let $\rho$ be the marginal probability of membership of $x$ in the positive class and $1 - \rho$ the marginal probability of membership in the negative class. These marginal probabilities sometimes are referred to as the "class priors" or the "base rate."

A classifier $t$ is a mapping from instances $x$ to classes $\{p, n\}$ and is an approximation of $c$. For notational convenience, let $t(x) \in \{P, N\}$ so that it is always clear whether a class value is an actual (lower case) or predicted (upper case) value. The expected accuracy of a classifier $t$, $\alpha_t$, is defined as $\alpha_t = Pr(t(x) = c(x))$, or, equivalently as:

$$\alpha_t = \rho \bullet Pr(\text{t}(x) = \text{P} \mid \text{c}(x) = \text{p}) + (1 - \rho) \bullet Pr(\text{t}(x) = \text{N} \mid \text{c}(x) = \text{n}) \qquad [1]$$

Many classifiers produce not only a classification, but also estimates of the probability that $x$ will take on each class value. Let $\text{Post}_t(x)$ be classifier $t$'s estimated (posterior) probability that for instance $x$, $\text{c}(x) = \text{p}$. Classifiers that produce class-membership probabilities produce a classification by applying a numeric threshold to the posterior probabilities. For example, a threshold value of .5 may be used so that $t(x) = \text{P}$ *iff* $\text{Post}_t(x) > .5$; otherwise $t(x) = \text{N}$.

A variety of classifiers function by partitioning the input space into a set $L$ of disjoint regions (a region being defined by a set of potential instances). For example, for a classification tree, the regions are described by conjoining the conditions leading to the leaves of the tree. Each region $L \in L$ will contain some number of training instances, $\lambda_L$. Let $\lambda_{L_p}$ and $\lambda_{L_n}$ be the numbers of positive and negative training instances in region L, such that $\lambda_L = \lambda_{L_p} + \lambda_{L_n}$. Such classifiers often estimate $Post_t(x \mid x \in L)$ as $\lambda_{Lp}/(\lambda_{Lp} + \lambda_{Ln})$ and assign a classification for all instances $x \in L$ based on this estimate and a numeric threshold, as described earlier. Now, let $L_P$ and $L_N$ be the sets of regions that predict the positive and negative classes, respectively, such that $L_P \cup L_N = L$. For each region $L \in L$, $t$ has an associated accuracy, $\alpha_L = Pr(c(x) = t(x) \mid x \in L)$. Let $\alpha_{L_P}$ represent the expected accuracy for $x \in L_P$ and $\alpha_{L_N}$ the expected accuracy for $x \in L_N$.[2] As we shall see in Section 5.2, we expect $\alpha_{L_P} \neq \alpha_{L_N}$ when $\rho \neq .5$.

---

1. This paper addresses binary classification; the positive class always corresponds to the minority class and the negative class to the majority class.

## 3. Correcting for Changes to the Class Distribution of the Training Set

Many classifier induction algorithms assume that the training and test data are drawn from the same fixed, underlying, distribution $D$. In particular, these algorithms assume that $r_{train}$ and $r_{test}$, the fractions of positive examples in the training and test sets, approximate $\rho$, the true "prior" probability of encountering a positive example. These induction algorithms use the estimated class priors based on $r_{train}$, either implicitly or explicitly, to construct a model and to assign classifications. If the estimated value of the class priors is not accurate, then the posterior probabilities of the model will be improperly biased. Specifically, "increasing the prior probability of a class increases the posterior probability of the class, moving the classification boundary for that class so that more cases are classified into the class" (SAS, 2001). Thus, if the training-set data are selected so that $r_{train}$ does not approximate $\rho$, then the posterior probabilities should be adjusted based on the differences between $\rho$ and $r_{train}$. If such a correction is not performed, then the resulting bias will cause the classifier to classify the preferentially sampled class more accurately, but the overall accuracy of the classifier will almost always suffer (we discuss this further in Section 4 and provide the supporting evidence in Appendix A).[3]

In the majority of experiments described in this article the class distribution of the training set is purposefully altered so that $r_{train}$ does not approximate $\rho$. The purpose for modifying the class distribution of the training set is to evaluate how this change affects the overall performance of the classifier—and whether it can produce better-performing classifiers. However, we do not want the biased posterior probability estimates to affect the results. In this section we describe a method for adjusting the posterior probabilities to account for the difference between $r_{train}$ and $\rho$. This method (Weiss & Provost, 2001) is justified informally, using a simple, intuitive, argument. Elkan (2001) presents an equivalent method for adjusting the posterior probabilities, including a formal derivation.

When learning classification trees, differences between $r_{train}$ and $\rho$ normally result in biased posterior class-probability estimates at the leaves. To remove this bias, we adjust the probability estimates to take these differences into account. Two simple, common probability estimation formulas are listed in Table 1. For each, let $\lambda_{Lp}$ ($\lambda_{Ln}$) represent the number of minority-class (majority-class) training examples at a leaf L of a decision tree (or, more generally, within any region L). The uncorrected estimates, which are based on the assumption that the training and test sets are drawn from $D$ and approximate $\rho$, estimate the probability of seeing a minority-class (positive) example in L. The uncorrected frequency-based estimate is straightforward and requires no explanation. However, this estimate does not perform well when the sample size, $\lambda_{Lp} + \lambda_{Ln}$, is small—and is not even defined when the sample size is 0. For these reasons the

---

2. For notational convenience we treat $L_P$ and $L_N$ as the union of the sets of instances in the corresponding regions.

3. In situations where it is more costly to misclassify minority-class examples than majority-class examples, practitioners sometimes introduce this bias on purpose.

Laplace estimate often is used instead. We consider a version based on the Laplace law of succession (Good, 1965). This probability estimate will always be closer to 0.5 than the frequency-based estimate, but the difference between the two estimates will be negligible for large sample sizes.

| Estimate Name | Uncorrected | Corrected |
|---|---|---|
| Frequency-Based | $\lambda_{Lp}/(\lambda_{Lp}+\lambda_{Ln})$ | $\lambda_{Lp}/(\lambda_{Lp}+o\,\lambda_{Ln})$ |
| Laplace (law of succession) | $(\lambda_{Lp}+1)/(\lambda_{Lp}+\lambda_{Ln}+2)$ | $(\lambda_{Lp}+1)/(\lambda_{Lp}+o\,\lambda_{Ln}+2)$ |

Table 1: Probability Estimates for Observing a Minority-Class Example

The corrected versions of the estimates in Table 1 account for differences between $r_{train}$ and $\rho$ by factoring in the over-sampling ratio $o$, which measures the degree to which the minority class is over-sampled in the training set relative to the naturally occurring distribution. The value of $o$ is computed as the ratio of minority-class examples to majority-class examples in the training set divided by the same ratio in the naturally occurring class distribution. If the ratio of minority to majority examples were 1:2 in the training set and 1:6 in the naturally occurring distribution, then $o$ would be 3. A learner can account properly for differences between $r_{train}$ and $\rho$ by using the corrected estimates to calculate the posterior probabilities at L.

As an example, if the ratio of minority-class examples to majority-class examples in the naturally occurring class distribution is 1:5 but the training distribution is modified so that the ratio is 1:1, then $o$ is 1.0/0.2, or 5. For L to be labeled with the minority class the probability must be greater than 0.5, so, using the corrected frequency-based estimate, $\lambda_{Lp}/(\lambda_{Lp}+5\lambda_{Ln}) > 0.5$, or, $\lambda_{Lp} > 5\,\lambda_{Ln}$. Thus, L is labeled with the minority class only if it covers $o$ times as many minority-class examples as majority-class examples. Note that in calculating $o$ above we use the class ratios and not the fraction of examples belonging to the minority class (if we mistakenly used the latter in the above example, then $o$ would be one-half divided by one-sixth, or 3). Using the class ratios substantially simplifies the formulas and leads to more easily understood estimates. Elkan (2001) provides a more complex, but equivalent, formula that uses fractions instead of ratios. In this discussion we assume that a good approximation of the true base rate is known. In some real-world situations this is not true and different methods are required to compensate for changes to the training set (Provost & Fawcett, 2001; Saerens et al., 2002).

In order to demonstrate the importance of using the corrected estimates, Appendix A presents results comparing decision trees labeled using the uncorrected frequency-based estimate with trees using the corrected frequency-based estimate. This comparison shows that for a particular modification of the class distribution of the training sets (they are modified so that the classes are balanced), using the corrected estimates yields classifiers that substantially outperform classifiers labeled using the uncorrected estimate. In particular, over the twenty-six data sets used in our study, the corrected frequency-based estimate yields a relative reduction in error rate of 10.6%. Furthermore, for only one of the twenty-six data sets does the corrected estimate perform worse. Consequently it is critical to take the differences in the class distributions into account when labeling the leaves. Previous work on modifying the class distribution of the training set (Catlett, 1991; Chan & Stolfo, 1998; Japkowicz, 2002) did not take these differences into account and this undoubtedly affected the results.

## 4. Experimental Setup

In this section we describe the data sets analyzed in this article, the sampling strategy used to alter the class distribution of the training data, the classifier induction program used, and, finally, the metrics for evaluating the performance of the induced classifiers.

### 4.1 The Data Sets and the Method for Generating the Training Data

The twenty-six data sets used throughout this article are described in Table 2. This collection includes twenty data sets from the UCI repository (Blake & Merz, 1998), five data sets, identified with a "+", from previously published work by researchers at AT&T (Cohen & Singer, 1999), and one new data set, the phone data set, generated by the authors. The data sets in Table 2 are listed in order of decreasing class imbalance, a convention used throughout this article.

| Dataset | % Minority Examples | Dataset Size | # | Dataset | % Minority Examples | Dataset Size |
|---|---|---|---|---|---|---|
| letter-a* | 3.9 | 20,000 | 14 | network2 | 27.9 | 3,826 |
| pendigits* | 8.3 | 13,821 | 15 | yeast* | 28.9 | 1,484 |
| abalone* | 8.7 | 4,177 | 16 | network1+ | 29.2 | 3,577 |
| sick-euthyroid | 9.3 | 3,163 | 17 | car* | 30.0 | 1,728 |
| connect-4* | 9.5 | 11,258 | 18 | german | 30.0 | 1,000 |
| optdigits* | 9.9 | 5,620 | 19 | breast-wisc | 34.5 | 699 |
| covertype* | 14.8 | 581,102 | 20 | blackjack+ | 35.6 | 15,000 |
| solar-flare* | 15.7 | 1,389 | 21 | weather+ | 40.1 | 5,597 |
| phone | 18.2 | 652,557 | 22 | bands | 42.2 | 538 |
| letter-vowel* | 19.4 | 20,000 | 23 | market1+ | 43.0 | 3,181 |
| contraceptive* | 22.6 | 1,473 | 24 | crx | 44.5 | 690 |
| adult | 23.9 | 48,842 | 25 | kr-vs-kp | 47.8 | 3,196 |
| splice-junction* | 24.1 | 3,175 | 26 | move+ | 49.9 | 3,029 |

Table 2: Description of Data Sets

In order to simplify the presentation and the analysis of our results, data sets with more than two classes were mapped to two-class problems. This was accomplished by designating one of the original classes, typically the least frequently occurring class, as the minority class and then mapping the remaining classes into the majority class. The data sets that originally contained more than 2 classes are identified with an asterisk (*) in Table 2. The letter-a data set was created from the letter-recognition data set by assigning the examples labeled with the letter "a" to the minority class; the letter-vowel data set was created by assigning the examples labeled with any vowel to the minority class.

We generated training sets with different class distributions as follows. For each experimental run, first the test set is formed by randomly selecting 25% of the minority-class examples and 25% of the majority-class examples from the original data set, without replacement (the resulting test set therefore conforms to the original class distribution). The remaining data are available for training. To ensure that all experiments for a given data set have the same training-set size—no matter what the class distribution of the training set—the training-set size, S, is made equal to the total number of minority-class examples still available for training (i.e., 75% of the original number). This makes it possible, without replicating any examples, to generate any class distribution for training-set size S. Each training set is then formed by random sampling from the remaining data, without replacement, such that the desired class distribution is achieved. For the experiments described in this article, the class distribution of the training set is varied so that the minority class accounts for between 2% and 95% of the training data.

## 4.2 C4.5 and Pruning

The experiments in this article use C4.5, a program for inducing classification trees from labeled examples (Quinlan, 1993). C4.5 uses the uncorrected frequency-based estimate to label the leaves of the decision tree, since it assumes that the training data approximate the true, underlying distribution. Given that we modify the class distribution of the training set, it is essential that we use the corrected estimates to re-label the leaves of the induced tree. The results presented in the body of this article are based on the use of the corrected versions of the frequency-based and Laplace estimates (described in Table 1), using a probability threshold of .5 to label the leaves of the induced decision trees.

C4.5 does not factor in differences between the class distributions of the training and test sets—we adjust for this as a post-processing step. If C4.5's pruning strategy, which attempts to minimize error rate, were allowed to execute, it would prune based on a false assumption (viz., that the test distribution matches the training distribution). Since this may negatively affect the generated classifier, except where otherwise indicated all results are based on C4.5 without pruning. This decision is supported by recent research, which indicates that when target misclassification costs (or class distributions) are unknown then standard pruning does not improve, and may degrade, generalization performance (Provost & Domingos, 2001; Zadrozny & Elkan, 2001; Bradford et al., 1998; Bauer & Kohavi, 1999). Indeed, Bradford et al. (1998) found that even if the pruning strategy is adapted to take misclassification costs and class distribution into account, this does not generally improve the performance of the classifier. Nonetheless, in order to justify using C4.5 without pruning, we also present the results of C4.5 with pruning when the training set uses the natural distribution. In this situation C4.5's assumption about $r_{train}$ approximating $\rho$ is valid and hence its pruning strategy will perform properly. Looking ahead, these results show that C4.5 without pruning indeed performs competitively with C4.5 with pruning.

## 4.3 Evaluating Classifier Performance

A variety of metrics for assessing classifier performance are based on the terms listed in the confusion matrix shown below.

|  |  | $t(x)$ | |
|---|---|---|---|
|  |  | Positive Prediction | Negative Prediction |
| $c(x)$ | Actual Positive | $tp$ (true positive) | $fn$ (false negative) |
|  | Actual Negative | $fp$ (false positive) | $tn$ (true negative) |

Table 3 summarizes eight such metrics. The metrics described in the first two rows measure the ability of a classifier to classify positive and negative *examples* correctly, while the metrics described in the last two rows measure the effectiveness of the *predictions* made by a classifier. For example, the positive predictive value (PPV), or *precision*, of a classifier measures the fraction of positive predictions that are correctly classified. The metrics described in the last two rows of Table 3 are used throughout this article to evaluate how various training-set class distributions affect the predictions made by the induced classifiers. Finally, the metrics in the second column of Table 3 are "complements" of the corresponding metrics in the first column, and can alternatively be computed by subtracting the value in the first column from 1. More specifically, proceeding from row 1 through 4, the metrics in column 1 (column 2) represent: 1) the accuracy (error rate) when classifying positive/minority examples, 2) the accuracy (error rate) when classifying negative/minority examples, 3) the accuracy (error rate) of the positive/minority predictions, and 4) the accuracy (error rate) of the negative/majority predictions.

| | | | | |
|---|---|---|---|---|
| $TP = Pr(P|p) \approx \dfrac{tp}{tp+fn}$ | *True Positive Rate (recall or sensitivity)* | $FN = Pr(N|p) \approx \dfrac{fn}{tp+fn}$ | *False Negative Rate* |
| $TN = Pr(N|n) \approx \dfrac{tn}{tn+fp}$ | *True Negative Rate (specificity)* | $FP = Pr(P|n) \approx \dfrac{fp}{tn+fp}$ | *False Positive Rate* |
| $PPV = Pr(p|P) \approx \dfrac{tp}{tp+fp}$ | *Positive Predictive Value (precision)* | $\overline{PPV} = Pr(n|P) \approx \dfrac{fp}{tp+fp}$ | |
| $NPV = Pr(n|N) \approx \dfrac{tn}{tn+fn}$ | *Negative Predictive Value* | $\overline{NPV} = Pr(y|N) \approx \dfrac{fn}{tn+fn}$ | |

Table 3: Classifier Performance Metrics

We use two performance measures to gauge the *overall* performance of a classifier: classification accuracy and the area under the ROC curve (Bradley, 1997). Classification accuracy is $(tp + fp)/(tp + fp + tn + fn)$. This formula, which represents the fraction of examples that are correctly classified, is an estimate of the expected accuracy, $\alpha_t$, defined earlier in equation 1. Throughout this article we specify classification accuracy in terms of error rate, which is $1 - accuracy$.

We consider classification accuracy in part because it is the most common evaluation metric in machine-learning research. However, using accuracy as a performance measure assumes that the target (marginal) class distribution is known and unchanging and, more importantly, that the error costs—the costs of a false positive and false negative—are equal. These assumptions are unrealistic in many domains (Provost et al., 1998). Furthermore, highly unbalanced data sets typically have highly non-uniform error costs that favor the minority class, which, as in the case of medical diagnosis and fraud detection, is the class of primary interest. The use of accuracy in these cases is particularly suspect since, as we discuss in Section 5.2, it is heavily biased to favor the majority class and therefore will sometimes generate classifiers that never predict the minority class. In such cases, Receiver Operating Characteristic (ROC) analysis is more appropriate (Swets et al., 2000; Bradley, 1997; Provost & Fawcett, 2001). When producing the ROC curves we use the Laplace estimate to estimate the probabilities at the leaves, since it has been shown to yield consistent improvements (Provost & Domingos, 2001). To assess the *overall* quality of a classifier we measure the fraction of the total area that falls under the ROC curve (AUC), which is equivalent to several other statistical measures for evaluating classification and ranking models (Hand, 1997). Larger AUC values indicate generally better classifier performance and, in particular, indicate a better ability to rank cases by likelihood of class membership.

## 5. Learning from Unbalanced Data Sets

We now analyze the classifiers induced from the twenty-six naturally unbalanced data sets described in Table 2, focusing on the differences in performance for the minority and majority classes. We do not alter the class distribution of the training data in this section, so the classifiers need not be adjusted using the method described in Section 3. However, so that these experiments are consistent with those in Section 6 that use the natural distribution, the size of the training set is reduced, as described in Section 4.1.

Before addressing these differences, it is important to discuss an issue that may lead to confusion if left untreated. Practitioners have noted that learning performance often is unsatisfactory when learning from data sets where the minority class is substantially underrepresented. In particular, they observe that there is a large error rate for the minority class. As should be clear from Table 3 and the associated discussion, there are two different notions of "error rate for the minority class": the minority-class predictions could have a high error rate (large $\overline{PPV}$) or the minority-class test examples could have a high error rate (large $FN$). When practitioners observe that the error rate is unsatisfactory for the minority class, they are usually referring to the fact that the minority-class *examples* have a high error rate (large $FN$). The analysis in this section will show that the error rate associated with the minority-class predictions ($\overline{PPV}$) and the minority-class test examples ($FN$) both are much larger than their majority-class counterparts ($\overline{NPV}$ and $FP$, respectively). We discuss several explanations for these observed differences.

### 5.1 Experimental Results

The performances of the classifiers induced from the twenty-six unbalanced data sets are described in Table 4. This table warrants some explanation. The first column specifies the data set name while the second column, which for convenience has been copied from Table 2, specifies the percentage of minority-class examples in the natural class distribution. The third column specifies the percentage of the total test errors that can be attributed to the test examples that belong to the minority class. By comparing the values in columns two and three we see that in all cases a disproportionately large percentage of the errors come from the minority-class examples. For instance, minority-class examples make up only 3.9% of the letter-a data set but contribute 58.3% of the errors. Furthermore, for 22 of 26 data sets a *majority* of the errors can be attributed to minority-class examples.

The fourth column specifies the number of leaves labeled with the minority and majority classes and shows that in all but two cases there are fewer leaves labeled with the minority class than with the majority class. The fifth column, "Coverage," specifies the average number of training examples that each minority-labeled or majority-labeled leaf classifies ("covers"). These results indicate that the leaves labeled with the minority class are formed from far fewer training examples than those labeled with the majority class.

The "Prediction ER" column specifies the error rates associated with the minority-class and majority-class predictions, based on the performance of these predictions at classifying the test examples. The "Actuals ER" column specifies the classification error rates for the minority and majority class examples, again based on the test set. These last two columns are also labeled using the terms defined in Section 2 ($\overline{PPV}$, $\overline{NPV}$, FN, and FP). As an example, these columns show that for the letter-a data set the minority-labeled predictions have an error rate of 32.5% while the majority-labeled predictions have an error rate of only 1.7%, and that the minority-class test examples have a classification error rate of 41.5% while the majority-class test exam-ples have an error rate of only 1.2%. In each of the last two columns we underline the higher error rate.

| Dataset | % Minority Examples | % Errors from Min. | Leaves | | Coverage | | Prediction ER | | Actuals ER | |
|---|---|---|---|---|---|---|---|---|---|---|
| | | | Min. | Maj. | Min. | Maj. | Min. (PPV) | Maj. (NPV) | Min. (FN) | Maj. (FP) |
| letter-a | 3.9 | 58.3 | 11 | 138 | 2.2 | 4.3 | 32.5 | 1.7 | 41.5 | 1.2 |
| pendigits | 8.3 | 32.4 | 6 | 8 | 16.8 | 109.3 | 25.8 | 1.3 | 14.3 | 2.7 |
| abalone | 8.7 | 68.9 | 5 | 8 | 2.8 | 35.5 | 69.8 | 7.7 | 84.4 | 3.6 |
| sick-euthyroid | 9.3 | 51.2 | 4 | 9 | 7.1 | 26.9 | 22.5 | 2.5 | 24.7 | 2.4 |
| connect-4 | 9.5 | 51.4 | 47 | 128 | 1.7 | 5.8 | 55.8 | 6.0 | 57.6 | 5.7 |
| optdigits | 9.9 | 73.0 | 15 | 173 | 2.9 | 2.4 | 18.0 | 3.9 | 36.7 | 1.5 |
| covertype | 14.8 | 16.7 | 350 | 446 | 27.3 | 123.2 | 23.1 | 1.0 | 5.7 | 4.9 |
| solar-flare | 15.7 | 64.4 | 12 | 48 | 1.7 | 3.1 | 67.8 | 13.7 | 78.9 | 8.1 |
| phone | 18.2 | 64.4 | 1008 | 1220 | 13.0 | 62.7 | 30.8 | 9.5 | 44.6 | 5.5 |
| letter-vowel | 19.4 | 61.8 | 233 | 2547 | 2.4 | 0.9 | 27.0 | 8.7 | 37.5 | 5.6 |
| contraceptive | 22.6 | 48.7 | 31 | 70 | 1.8 | 2.8 | 69.8 | 20.1 | 68.3 | 21.1 |
| adult | 23.9 | 57.5 | 627 | 4118 | 3.1 | 1.6 | 34.3 | 12.6 | 41.5 | 9.6 |
| splice-junction | 24.1 | 58.9 | 26 | 46 | 5.5 | 9.6 | 15.1 | 6.3 | 20.3 | 4.5 |
| network2 | 27.9 | 57.1 | 50 | 61 | 4.0 | 10.3 | 48.2 | 20.4 | 55.5 | 16.2 |
| yeast | 28.9 | 58.9 | 8 | 12 | 14.4 | 26.1 | 45.6 | 20.9 | 55.0 | 15.6 |
| network1 | 29.2 | 57.1 | 42 | 49 | 5.1 | 12.8 | 46.2 | 21.0 | 53.9 | 16.7 |
| car | 30.0 | 58.6 | 38 | 42 | 3.1 | 6.6 | 14.0 | 7.7 | 18.6 | 5.6 |
| german | 30.0 | 55.4 | 34 | 81 | 2.0 | 2.0 | 57.1 | 25.4 | 62.4 | 21.5 |
| breast-wisc | 34.5 | 45.7 | 5 | 5 | 12.6 | 26.0 | 11.4 | 5.1 | 9.8 | 6.1 |
| blackjack | 35.6 | 81.5 | 13 | 19 | 57.7 | 188.0 | 28.9 | 27.9 | 64.4 | 8.1 |
| weather | 40.1 | 50.7 | 134 | 142 | 5.0 | 7.2 | 41.0 | 27.7 | 41.7 | 27.1 |
| bands | 42.2 | 91.2 | 52 | 389 | 1.4 | 0.3 | 17.8 | 34.8 | 69.8 | 4.9 |
| market1 | 43.0 | 50.3 | 87 | 227 | 5.1 | 2.7 | 30.9 | 23.4 | 31.2 | 23.3 |
| crx | 44.5 | 51.0 | 28 | 65 | 3.9 | 2.1 | 23.2 | 18.9 | 24.1 | 18.5 |
| kr-vs-kp | 47.8 | 54.0 | 23 | 15 | 24.0 | 41.2 | 1.2 | 1.3 | 1.4 | 1.1 |
| move | 49.9 | 61.4 | 235 | 1025 | 2.4 | 0.6 | 24.4 | 29.9 | 33.9 | 21.2 |
| Average | 25.8 | 56.9 | 120 | 426 | 8.8 | 27.4 | 33.9 | 13.8 | 41.4 | 10.1 |
| Median | 26.0 | 57.3 | 33 | 67 | 3.9 | 6.9 | 29.9 | 11.1 | 41.5 | 5.9 |

Table 4: Behavior of Classifiers Induced from Unbalanced Data Sets

The results in Table 4 clearly demonstrate that the minority-class predictions perform much worse than the majority-class predictions and that the minority-class examples are misclassified much more frequently than majority-class examples. Over the twenty-six data sets, the minority predictions have an average error rate ($\overline{PPV}$) of 33.9% while the majority-class predictions have an average error rate ($\overline{NPV}$) of only 13.8%. Furthermore, for only three of the twenty-six data sets do the majority-class predictions have a higher error rate—and for these three data sets the class distributions are only slightly unbalanced. Table 4 also shows us that the average error rate for the minority-class test examples (FN) is 41.4% whereas for the majority-class test examples the error rate (FP) is only 10.1%. In every one of the twenty-six cases the minority-class test examples have a higher error rate than the majority-class test examples.

### 5.2 Discussion

Why do the minority-class predictions have a higher error rate ($\overline{\text{PPV}}$) than the majority-class predictions ($\overline{\text{NPV}}$)? There are at least two reasons. First, consider a classifier $t_{\text{random}}$ where the partitions $L$ are chosen randomly and the assignment of each $L \in L$ to $L_{\text{P}}$ and $L_{\text{N}}$ is also made randomly (recall that $L_{\text{P}}$ and $L_{\text{N}}$ represent the regions labeled with the positive and negative classes). For a two-class learning problem the expected overall accuracy, $\alpha_t$, of this randomly generated and labeled classifier must be 0.5. However, the expected accuracy of the regions in the positive partition, $\alpha_{L_{\text{P}}}$, will be $\rho$ while the expected accuracy of the regions in the negative partition, $\alpha_{L_{\text{N}}}$, will be $1 - \rho$. For a highly unbalanced class distribution where $\rho = .01$, $\alpha_{L_{\text{P}}} = .01$ and $\alpha_{L_{\text{N}}} = .99$. Thus, in such a scenario the negative/majority predictions will be much more "accurate." While this "test distribution effect" will be small for a well-learned concept with a low Bayes error rate (and non-existent for a perfectly learned concept with a Bayes error rate of 0), many learning problems are quite hard and have high Bayes error rates.[4]

The results in Table 4 suggest a second explanation for why the minority-class predictions are so error prone. According to the coverage results, minority-labeled predictions tend to be formed from fewer training examples than majority-labeled predictions. *Small disjuncts*, which are the components of disjunctive concepts (i.e., classification rules, decision-tree leaves, etc.) that cover few training examples, have been shown to have a much higher error rate than large disjuncts (Holte, et al., 1989; Weiss & Hirsh, 2000). Consequently, the rules/leaves labeled with the minority class have a higher error rate partly because they suffer more from this "problem of small disjuncts."

Next, why are minority-class examples classified incorrectly much more often than majority-class examples (FN > FP)—a phenomenon that has also been observed by others (Japkowicz & Stephen, 2002)? Consider the estimated accuracy, $a_t$, of a classifier $t$, where the test set is drawn from the true, underlying distribution $D$:

$$a_t = \text{TP} \bullet r_{\text{test}} + \text{TN} \bullet (1 - r_{\text{test}}) \qquad [2]$$

Since the positive class corresponds to the minority class, $r_{\text{test}} < .5$, and for highly unbalanced data sets $r_{\text{test}} \ll .5$. Therefore, false-positive errors are more damaging to classification accuracy than false negative errors are. A classifier that is induced using an induction algorithm geared toward maximizing accuracy therefore should "prefer" false-negative errors over false-positive errors. This will cause negative/majority examples to be predicted more often and hence will lead to a higher error rate for minority-class examples. One straightforward example of how learning algorithms exhibit this behavior is provided by the common-sense rule: if there is no evidence favoring one classification over another, then predict the majority class. More generally, induction algorithms that maximize accuracy should be biased to perform better at classifying majority-class examples than minority-class examples, since the former component is weighted more heavily when calculating accuracy. This also explains why, when learning from data sets with a high degree of class imbalance, classifiers rarely predict the minority class.

A second reason why minority-class examples are misclassified more often than majority-class examples is that fewer minority-class examples are likely to be sampled from the distribu-tion $D$. Therefore, the training data are less likely to include (enough) instances of all of the minority-class subconcepts in the concept space, and the learner may not have the opportunity to represent all truly positive regions in $L_P$. Because of this, some minority-class test examples will be mistakenly classified as belonging to the majority class.

---

4. The (optimal) Bayes error rate, using the terminology from Section 2, occurs when $t(.)=c(.)$. Because $c(.)$ may be probabilistic (e.g., when noise is present), the Bayes error rate for a well-learned concept may not always be low. The test distribution effect will be small when the concept is well learned *and* the Bayes error rate is low.

Finally, it is worth noting that $\overline{PPV} > \overline{NPV}$ does not imply that FN > FP. That is, having more error-prone minority predictions does *not* imply that the minority-class examples will be misclassified more often than majority-class examples. Indeed, a higher error rate for minority predictions means more majority-class test examples will be misclassified. The reason we generally observe a lower error rate for the majority-class test examples (FN > FP) is because the majority class is predicted far more often than the minority class.

## 6. The Effect of Training-Set Class Distribution on Classifier Performance

We now turn to the central questions of our study: how do different training-set class distributions affect the performance of the induced classifiers and which class distributions lead to the best classifiers? We begin by describing the methodology for determining which class distribution performs best. Then, in the next two sections, we evaluate and analyze classifier performance for the twenty-six data sets using a variety of class distributions. We use error rate as the performance metric in Section 6.2 and AUC as the performance metric in Section 6.3.

### 6.1 Methodology for Determining the Optimum Training Class Distribution(s)

In order to evaluate the effect of class distribution on classifier performance, we vary the training-set class distributions for the twenty-six data sets using the methodology described in Section 4.1. We evaluate the following twelve class distributions (expressed as the percentage of minority-class examples): 2%, 5%, 10%, 20%, 30%, 40%, 50%, 60%, 70%, 80%, 90%, and 95%. For each data set we also evaluate the performance using the naturally occurring class distribution.

Before we try to determine the "best" class distribution for a training set, there are several issues that must be addressed. First, because we do not evaluate every possible class distribution, we can only determine the best distribution among the 13 evaluated distributions. Beyond this concern, however, is the issue of statistical significance and, because we generate classifiers for 13 training distributions, the issue of multiple comparisons (Jensen & Cohen, 2000). Because of these issues we cannot always conclude that the distribution that yields the best performing classifiers is truly the best one for training.

We take several steps to address the issues of statistical significance and multiple comparisons. To enhance our ability to identify true differences in classifier performance with respect to changes in class distribution, all results presented in this section are based on 30 runs, rather than the 10 runs employed in Section 5. Also, rather than trying to determine *the* best class distribution, we adopt a more conservative approach, and instead identify an "optimal range" of class distributions—a range in which we are confident the best distribution lies. To identify the optimal range of class distributions, we begin by identifying, for each data set, the class distribution that yields the classifiers that perform best over the 30 runs. We then perform t-tests to compare the performance of these 30 classifiers with the 30 classifiers generated using each of the other twelve class distributions (i.e., 12 t-tests each with n=30 data points). If a t-test yields a probability ≤ .10 then we conclude that the "best" distribution is different from the "other" distribution (i.e., we are at least 90% confident of this); otherwise we cannot conclude that the class distributions truly perform differently and therefore "group" the distributions together. These grouped distributions collectively form the "optimal range" of class distributions. As Tables 5 and 6 will show, in 50 of 52 cases the optimal ranges are contiguous, assuaging concerns that our conclusions are due to problems of multiple comparisons.

## 6.2 The Relationship between Class Distribution and Classification Error Rate

Table 5 displays the error rates of the classifiers induced for each of the twenty-six data sets. The first column in Table 5 specifies the name of the data set and the next two columns specify the error rates that result from using the natural distribution, with and then without pruning. The next 12 columns present the error rate values for the 12 fixed class distributions (without pruning). For each data set, the "best" distribution (i.e., the one with the lowest error rate) is highlighted by underlining it and displaying it in boldface. The relative position of the natural distribution within the range of evaluated class distributions is denoted by the use of a vertical bar between columns. For example, for the letter-a data set the vertical bar indicates that the natural distribution falls between the 2% and 5% distributions (from Table 2 we see it is 3.9%).

| Dataset | Error Rate when using Specified Training Distribution (training distribution expressed as % minority) | | | | | | | | | | | | | | Relative % Improvement | |
|---|---|---|---|---|---|---|---|---|---|---|---|---|---|---|---|---|
| | Nat-Prune | Nat | 2 | 5 | 10 | 20 | 30 | 40 | 50 | 60 | 70 | 80 | 90 | 95 | best vs. nat | best vs. bal |
| letter-a | 2.80 x | 2.78 | 2.86 | 2.75 | **2.59** | 3.03 | 3.79 | 4.53 | 5.38 | 6.48 | 8.51 | 12.37 | 18.10 | 26.14 | 6.8 | 51.9 |
| pendigits | 3.65 + | 3.74 | 5.77 | 3.95 | 3.63 | **3.45** | 3.70 | 3.64 | 4.02 | 4.48 | 4.98 | 5.73 | 8.83 | 13.36 | 7.8 | 14.2 |
| abalone | 10.68 x | 10.46 | **9.04** | 9.61 | 10.64 | 13.19 | 15.33 | 20.76 | 22.97 | 24.09 | 26.44 | 27.70 | 27.73 | 33.91 | 13.6 | 60.6 |
| sick-euthyroid | 4.46 x | **4.10** | 5.78 | 4.82 | 4.69 | 4.25 | 5.79 | 6.54 | 6.85 | 9.73 | 12.89 | 17.28 | 28.84 | 40.34 | 0.0 | 40.1 |
| connect-4 | 10.68 x | 10.56 | **7.65** | 8.66 | 10.80 | 15.09 | 19.31 | 23.18 | 27.57 | 33.09 | 39.45 | 47.24 | 59.73 | 72.08 | 27.6 | 72.3 |
| optdigits | 4.94 x | 4.68 | 8.91 | 7.01 | 4.05 | 3.05 | 2.83 | **2.79** | 3.41 | 3.87 | 5.15 | 5.75 | 9.72 | 12.87 | 40.4 | 18.2 |
| covertype | 5.12 x | 5.03 | 5.54 | 5.04 | **5.00** | 5.26 | 5.64 | 5.95 | 6.46 | 7.23 | 8.50 | 10.18 | 13.03 | 16.27 | 0.6 | 22.6 |
| solar-flare | 19.16 + | 19.98 | **16.54** | 17.52 | 18.96 | 21.45 | 23.03 | 25.49 | 29.12 | 30.73 | 33.74 | 38.31 | 44.72 | 52.22 | 17.2 | 43.2 |
| phone | 12.63 x | 12.62 | 13.45 | 12.87 | **12.32** | 12.68 | 13.25 | 13.94 | 14.81 | 15.97 | 17.32 | 18.73 | 20.24 | 21.07 | 2.4 | 16.8 |
| letter-vowel | 11.76 x | **11.63** | 15.87 | 14.24 | 12.53 | 11.67 | 12.00 | 12.69 | 14.16 | 16.00 | 18.88 | 23.47 | 32.20 | 41.81 | 0.0 | 17.9 |
| contraceptive | 31.71 x | 30.47 | **24.09** | 24.57 | 25.94 | 30.03 | 32.43 | 35.45 | 39.65 | 43.20 | 47.57 | 54.44 | 62.31 | 67.07 | 20.9 | 39.2 |
| adult | 17.42 x | 17.25 | 18.47 | 17.26 | **16.85** | 17.09 | 17.78 | 18.85 | 20.05 | 21.79 | 24.08 | 27.11 | 33.00 | 39.75 | 2.3 | 16.0 |
| splice-junction | 8.30 + | 8.37 | 20.00 | 13.95 | 10.72 | 8.68 | 8.50 | **8.15** | 8.74 | 9.86 | 9.85 | 12.08 | 16.25 | 21.18 | 2.6 | 6.8 |
| network2 | 27.13 x | 26.67 | 27.37 | 25.91 | 25.71 | **25.66** | 26.94 | 28.65 | 29.96 | 32.27 | 34.25 | 37.73 | 40.76 | 37.72 | 3.8 | 14.4 |
| yeast | 26.98 x | 26.59 | 29.08 | 28.61 | 27.51 | **26.35** | 26.93 | 27.10 | 28.80 | 29.82 | 30.91 | 35.42 | 35.79 | 36.33 | 0.9 | 8.5 |
| network1 | 27.57 + | 27.59 | 27.90 | 27.43 | 26.78 | **26.58** | 27.45 | 28.61 | 30.99 | 32.65 | 34.26 | 37.30 | 39.39 | 41.09 | 3.7 | 14.2 |
| car | 9.51 x | 8.85 | 23.22 | 18.58 | 14.90 | 10.94 | 8.63 | 8.31 | 7.92 | **7.35** | 7.79 | 8.78 | 10.18 | 12.86 | 16.9 | 7.2 |
| german | 33.76 x | 33.41 | **30.17** | 30.39 | 31.01 | 32.59 | 33.08 | 34.15 | 37.09 | 40.55 | 44.04 | 48.36 | 55.07 | 60.99 | 9.7 | 18.7 |
| breast-wisc | 7.41 x | 6.82 | 20.65 | 14.04 | 11.00 | 8.12 | 7.49 | 6.82 | **6.74** | 7.30 | 6.94 | 7.53 | 10.02 | 10.56 | 1.2 | 0.0 |
| blackjack | 28.14 x | **28.40** | 30.74 | 30.66 | 29.81 | 28.67 | 28.56 | 28.45 | 28.71 | 28.91 | 29.78 | 31.02 | 32.67 | 33.87 | 0.0 | 1.1 |
| weather | 33.68 + | 33.69 | 38.41 | 36.89 | 35.25 | 33.68 | **33.11** | 33.43 | 34.61 | 36.69 | 38.36 | 41.68 | 47.23 | 51.69 | 1.7 | 4.3 |
| bands | 32.26 + | 32.53 | 38.72 | 35.87 | 35.71 | 34.76 | 33.33 | **32.16** | 32.68 | 33.91 | 34.64 | 39.88 | 40.98 | 40.80 | 1.1 | 1.6 |
| market1 | 26.71 x | 26.16 | 34.26 | 32.50 | 29.54 | 26.95 | 26.13 | 26.05 | **25.77** | 26.86 | 29.53 | 31.69 | 36.72 | 39.90 | 1.5 | 0.0 |
| crx | 20.99 x | **20.39** | 35.99 | 30.86 | 27.68 | 23.61 | 20.84 | 20.82 | 21.48 | 21.64 | 22.20 | 23.98 | 28.09 | 32.85 | 0.0 | 5.1 |
| kr-vs-kp | 1.25 + | 1.39 | 12.18 | 6.50 | 3.20 | 2.33 | 1.73 | **1.16** | 1.22 | 1.34 | 1.53 | 2.55 | 3.66 | 6.04 | 16.5 | 4.9 |
| move | 27.54 + | 28.57 | 46.13 | 42.10 | 38.34 | 33.48 | 30.80 | 28.36 | **28.24** | 29.33 | 30.21 | 31.80 | 36.08 | 40.95 | 1.2 | 0.0 |

Table 5: Effect of Training Set Class Distribution on Error Rate

The error rate values that are not significantly different, statistically, from the lowest error rate (i.e., the comparison yields a t-test value > .10) are shaded. Thus, for the letter-a data set, the optimum range includes those class distributions that include between 2% and 10% minority-class examples—which includes the natural distribution. The last two columns in Table 5 show the relative improvement in error rate achieved by using the best distribution instead of the natural and balanced distributions. When this improvement is statistically significant (i.e., is associated with a t-test value ≤ .10) then the value is displayed in bold.

The results in Table 5 show that for 9 of the 26 data sets we are confident that the natural distribution is not within the optimal range. For most of these 9 data sets, using the best distribution rather than the natural distribution yields a remarkably large relative reduction in error rate. We feel that this is sufficient evidence to conclude that for accuracy, when the training-set size must be limited, it is not appropriate simply to assume that the natural distribution should be used. Inspection of the error-rate results in Table 5 also shows that the best distribution does not differ from the natural distribution in any consistent manner—sometimes it includes more minority-class examples (e.g., optdigits, car) and sometimes fewer (e.g., connect-4, solar-flare). However, it is clear that for data sets with a substantial amount of class imbalance (the ones in the top half of the table), a balanced class distribution also is not the best class distribution for training, to minimize undifferentiated error rate. More specifically, none of the top-12 most skewed data sets have the balanced class distribution within their respective optimal ranges, and for these data sets the relative improvements over the balanced distributions are striking.

Let us now consider the error-rate values for the remaining 17 data sets for which the t-test results do not permit us to conclude that the best observed distribution truly outperforms the natural distribution. In these cases we see that the error rate values for the 12 training-set class distributions usually form a unimodal, or nearly unimodal, distribution. This is the distribution one would expect if the accuracy of a classifier progressively degrades the further it deviates from the best distribution. This suggests that "adjacent" class distributions may indeed produce classifiers that perform differently, but that our statistical testing is not sufficiently sensitive to identify these differences. Based on this, we suspect that many of the observed improvements shown in the last column of Table 5 that are not deemed to be significant statistically are nonetheless meaningful.

Figure 1 shows the behavior of the learned classifiers for the adult, phone, covertype, and letter-a data sets in a graphical form. In this figure the natural distribution is denoted by the "X" tick mark and the associated error rate is noted above the marker. The error rate for the best distribution is underlined and displayed below the corresponding data point (for these four data sets the best distribution happens to include 10% minority-class examples). Two of the curves are associated with data sets (adult, phone) for which we are >90% confident that the best distribution performs better than the natural distribution, while for the other two curves (covertype, letter-a) we are not. Note that all four curves are perfectly unimodal. It is also clear that near the distribution that minimizes error rate, changes to the class distribution yield only modest changes in the error rate—far more dramatic changes occur elsewhere. This is also evident for most data sets in Table 5. This is a convenient property given the common goal of minimizing error rate. This property would be far less evident if the correction described in Section 3 were not performed, since then classifiers induced from class distributions deviating from the naturally occurring distribution would be improperly biased.

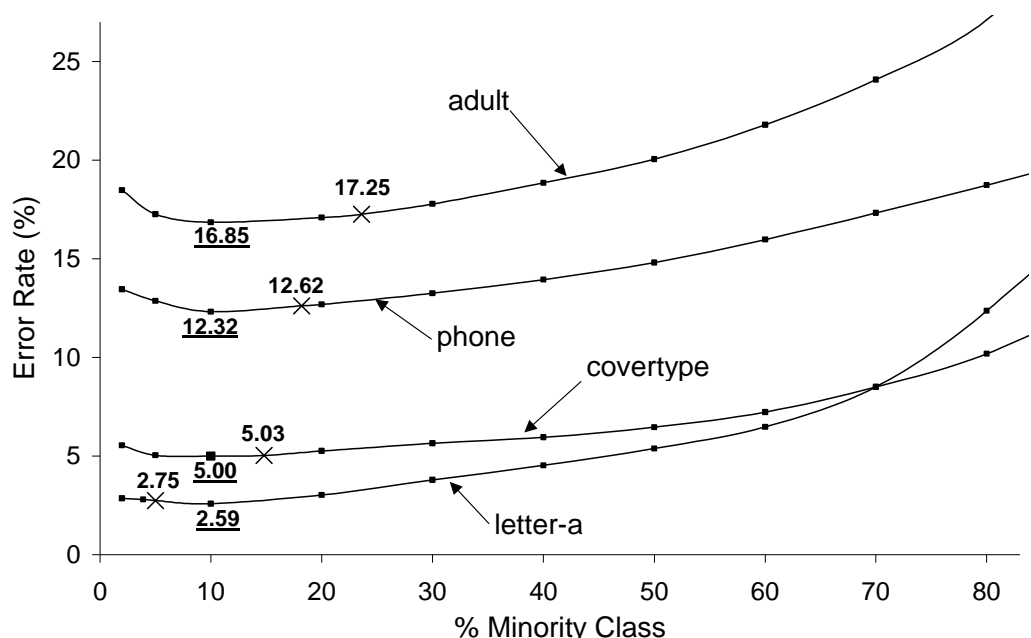

Figure 1: Effect of Class Distribution on Error Rate for Select Data Sets

Finally, to assess whether pruning would have improved performance, consider the second column in Table 5, which displays the error rates that result from using C4.5 with pruning on the natural distribution (recall from Section 4.2 that this is the only case when C4.5's pruning strategy will give unbiased results). A "+"/"x" in the second column indicates that C4.5 with pruning outperforms/underperforms C4.5 without pruning, when learning from the natural distribution. Note that C4.5 with pruning *underperforms* C4.5 without pruning for 17 of the 26 data sets, which leads us to conclude that C4.5 without pruning is a reasonable learner. Furthermore, in no case does C4.5 with pruning generate a classifier within the optimal range when C4.5 without pruning does not also generate a classifier within this range.

### 6.3 The Relationship between Class Distribution and AUC

The performance of the induced classifiers, using AUC as the performance measure, is displayed in Table 6. When viewing these results, recall that for AUC *larger* values indicate improved performance. The relative improvement in classifier performance is again specified in the last two columns, but now the relative improvement in performance is calculated in terms of the area above the ROC curve (i.e., 1 – AUC). We use the area above the ROC curve because it better reflects the relative improvement—just as in Table 5 relative improvement is specified in terms of the change in error rate instead of the change in accuracy. As before, the relative improvements are shown in bold only if we are more than 90% confident that they reflect a true improvement in performance (i.e., t-test value ≤ .10).

In general, the optimum ranges appear to be centered near, but slightly to the right, of the balanced class distribution. For 12 of the 26 data sets the optimum range does not include the natural distribution (i.e., the third column is not shaded). Note that for these data sets, with the exception of the solar-flare data set, the class distributions within the optimal range contain more minority-class examples than the natural class distribution. Based on these results we conclude even more strongly for AUC (i.e., for cost-sensitive classification and for ranking) than for accu-racy that it is not appropriate simply to choose the natural class distribution for training. Table 6 also shows that, unlike accuracy, a balanced class distribution generally performs very well, although it does not always perform optimally. In particular, we see that for 19 of the 26 data sets the balanced distribution is within the optimal range. This result is not too surprising since AUC, unlike error rate, is unaffected by the class distribution of the test set, and effectively factors in classifier performance over *all* class distributions.

| Dataset | AUC when using Specified Training Distribution (training distribution expressed as % minority) | | | | | | | | | | | | | | Relative % Improv. (1-AUC) | |
|---|---|---|---|---|---|---|---|---|---|---|---|---|---|---|---|---|
| | Nat-prune | Nat | 2 | 5 | 10 | 20 | 30 | 40 | 50 | 60 | 70 | 80 | 90 | 95 | best vs. nat | best vs. bal |
| letter-a | .500 x | .772 | .711 | .799 | .865 | .891 | .911 | .938 | .937 | .944 | .951 | **.954** | .952 | .940 | 79.8 | 27.0 |
| pendigits | .962 x | .967 | .892 | .958 | .971 | .976 | .978 | **.979** | **.979** | .978 | .977 | .976 | .966 | .957 | 36.4 | 0.0 |
| abalone | .590 x | .711 | .572 | .667 | .710 | .751 | .771 | .775 | .776 | **.778** | .768 | .733 | .694 | .687 | 25.8 | 0.9 |
| sick-euthyroid | .937 x | .940 | .892 | .908 | .933 | .943 | .944 | .949 | .952 | .951 | **.955** | .945 | .942 | .921 | 25.0 | 6.3 |
| connect-4 | .658 x | .731 | .664 | .702 | .724 | .759 | .763 | .777 | .783 | **.793** | **.793** | .789 | .772 | .730 | 23.1 | 4.6 |
| optdigits | .659 x | .803 | .599 | .653 | .833 | .900 | .924 | .943 | .948 | .959 | .967 | .965 | **.970** | .965 | 84.8 | 42.3 |
| covertype | .982 x | **.984** | .970 | .980 | **.984** | **.984** | .983 | .982 | .980 | .978 | .976 | .973 | .968 | .960 | 0.0 | 20.0 |
| solar-flare | .515 x | .627 | .614 | .611 | .646 | .627 | .635 | .636 | .632 | .650 | **.662** | .652 | .653 | .623 | 9.4 | 8.2 |
| phone | .850 x | .851 | .843 | .850 | .852 | .851 | .850 | .850 | .849 | .848 | .848 | .850 | **.853** | .850 | 1.3 | 2.6 |
| letter-vowel | .806 + | .793 | .635 | .673 | .744 | .799 | .819 | .842 | .849 | .861 | **.868** | **.868** | .858 | .833 | 36.2 | 12.6 |
| contraceptive | .539 x | .611 | .567 | .613 | .617 | .616 | .622 | .640 | .635 | .635 | .640 | **.641** | .627 | .613 | 7.7 | 1.6 |
| adult | .853 + | .839 | .816 | .821 | .829 | .836 | .842 | .846 | .851 | .854 | .858 | **.861** | **.861** | .855 | 13.7 | 6.7 |
| splice-junction | .932 + | .905 | .814 | .820 | .852 | .908 | .915 | .925 | .936 | .938 | .944 | **.950** | .944 | .944 | 47.4 | 21.9 |
| network2 | .712 + | .708 | .634 | .696 | .703 | .708 | .705 | .704 | .705 | .702 | .706 | .710 | **.719** | .683 | 3.8 | 4.7 |
| yeast | .702 x | .705 | .547 | .588 | .650 | .696 | **.727** | .714 | .720 | .723 | .715 | .699 | .659 | .621 | 10.9 | 2.5 |
| network1 | .707 + | .705 | .626 | .676 | .697 | .709 | .709 | .706 | .702 | .704 | .708 | **.713** | .709 | .696 | 2.7 | 3.7 |
| car | .931 + | .879 | .754 | .757 | .787 | .851 | .884 | .892 | .916 | .932 | .931 | **.936** | .930 | .915 | 47.1 | 23.8 |
| german | **.660** + | .646 | .573 | .600 | .632 | .615 | .635 | **.654** | .645 | .640 | .650 | .645 | .643 | .613 | 2.3 | 2.5 |
| breast-wisc | .951 x | .958 | .876 | .916 | .940 | .958 | .963 | **.968** | .966 | .963 | .963 | .964 | .949 | .948 | 23.8 | 5.9 |
| blackjack | .682 x | .700 | .593 | .596 | .628 | .678 | .688 | .712 | .713 | **.715** | .700 | .678 | .604 | .558 | 5.0 | 0.7 |
| weather | **.748** + | .736 | .694 | .715 | .728 | .737 | .738 | **.740** | .736 | .730 | .736 | .722 | .718 | .702 | 1.5 | 1.5 |
| bands | .604 x | **.623** | .522 | .559 | .564 | .575 | .599 | .620 | .618 | .604 | .601 | .530 | .526 | .536 | 0.0 | 1.3 |
| market1 | .815 + | .811 | .724 | .767 | .785 | .801 | .810 | .808 | .816 | **.817** | .812 | .805 | .795 | .781 | 3.2 | 0.5 |
| crx | **.889** + | .852 | .804 | .799 | .805 | .817 | .834 | .843 | .853 | .845 | .857 | .848 | .853 | **.866** | 9.5 | 8.8 |
| kr-vs-kp | .996 x | .997 | .937 | .970 | .991 | .994 | .997 | **.998** | **.998** | **.998** | .997 | .994 | .988 | .982 | 33.3 | 0.0 |
| move | **.762** + | .734 | .574 | .606 | .632 | .671 | .698 | .726 | .735 | .738 | **.742** | .736 | .711 | .672 | 3.0 | 2.6 |

Table 6: Effect of Training Set Class Distribution on AUC

If we look at the results with pruning, we see that for 15 of the 26 data sets C4.5 with pruning underperforms C4.5 without pruning. Thus, with respect to AUC, C4.5 without pruning is a reasonable learner. However, note that for the car data set the natural distribution with pruning falls into the optimum range, whereas without pruning it does not.

Figure 2 shows how class distribution affects AUC for the adult, covertype, and letter-a data sets (the phone data set is not displayed as it was in Figure 1 because it would obscure the adult data set). Again, the natural distribution is denoted by the "X" tick mark. The AUC for the best distribution is underlined and displayed below the corresponding data point. In this case we also see that near the optimal class distribution the AUC curves tend to be flatter, and hence less sensitive to changes in class distribution.

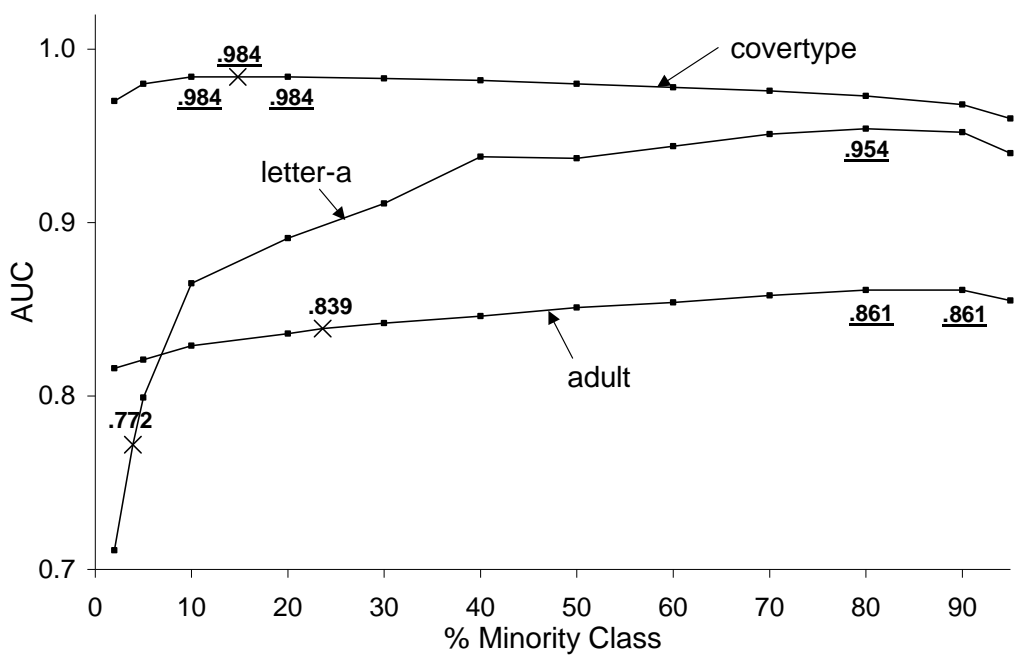

Figure 2: Effect of Class Distribution on AUC for Select Data Sets

Figure 3 shows several ROC curves associated with the letter-vowel data set. These curves each were generated from a single run of C4.5 (which is why the AUC values do not exactly match the values in Table 6). In ROC space, the point (0,0) corresponds to the strategy of never making a positive/minority prediction and the point (1,1) to always predicting the positive/minority class. Points to the "northwest" indicate improved performance.

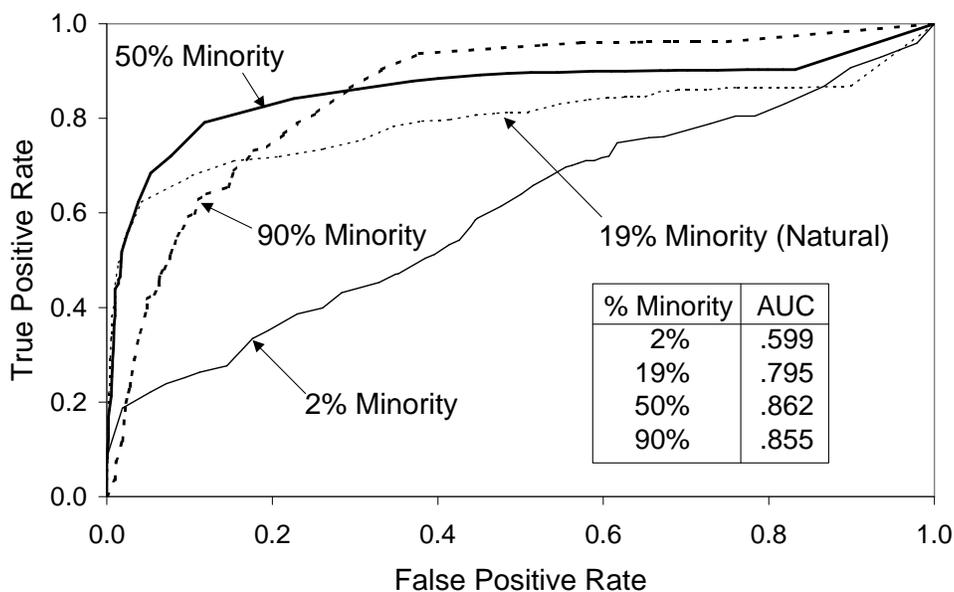

Figure 3: ROC Curves for the Letter-Vowel Data set

Observe that different training distributions perform better in different areas of ROC space. Specifically note that the classifier trained with 90% minority-class examples performs substantially better than the classifier trained with the natural distribution for high true-positive rates and that the classifier training with 2% minority-class examples performs fairly well for low true-positive rates. Why? With only a small sample of minority-class examples (2%) a classifier can identify only a few minority-labeled "rules" with high confidence. However, with a much larger sample of minority-class examples (90%) it can identify many more such minority-labeled rules. However, for this data set a balanced distribution has the largest AUC and performs best overall. Note that the curve generated using the balanced class distribution almost always outperforms the curve associated with the natural distribution (for low false-positive rates the natural distribution performs slightly better).

## 7. Forming a Good Class Distribution with Sensitivity to Procurement Costs

The results from the previous section demonstrate that some marginal class distributions yield classifiers that perform substantially better than the classifiers produced by other training distributions. Unfortunately, in order to determine the best class distribution for training, forming all thirteen training sets of size $n$, each with a different class distribution, requires nearly $2n$ examples. When it is costly to obtain training examples in a form suitable for learning, then this approach is self-defeating. Ideally, given a budget that allows for $n$ training examples, one would select a total of $n$ training examples all of which would be used in the final training set—and the associated class distribution would yield classifiers that perform better than those generated from any other class distribution (given $n$ training examples). In this section we describe and evaluate a heuristic, budget-sensitive, progressive sampling algorithm that approximates this ideal.

In order to evaluate this progressive sampling algorithm, it is necessary to measure how class distribution affects classifier performance for a variety of different training-set sizes. These measurements are summarized in Section 7.1 (the detailed results are included in Appendix B). The algorithm for selecting training data is then described in Section 7.2 and its performance evaluated in Section 7.3, using the measurements included in Appendix B.

### 7.1 The Effect of Class Distribution and Training-Set Size on Classifier Performance

Experiments were run to establish the relationship between class distribution, training-set size and classifier performance. In order to ensure that the training sets contain a sufficient number of training examples to provide meaningful results when the training-set size is dramatically reduced, only the data sets that yield relatively large training sets are used (this is determined based on the size of the data set and the fraction of minority-class examples in the data set). Based on this criterion, the following seven data sets were selected for analysis: phone, adult, covertype, blackjack, kr-vs-kp, letter-a, and weather. The detailed results associated with these experiments are contained in Appendix B.

The results for one of these data sets, the adult data set, are shown graphically in Figure 4 and Figure 5, which show classifier performance using error rate and AUC, respectively. Each of the nine performance curves in these figures is associated with a specific training-set size, which contains between 1/128 and all of the training data available for learning (using the methodology described in Section 4.1). Because the performance curves always improve with increasing data-set size, only the curves corresponding to the smallest and largest training-set sizes are explicitly labeled.

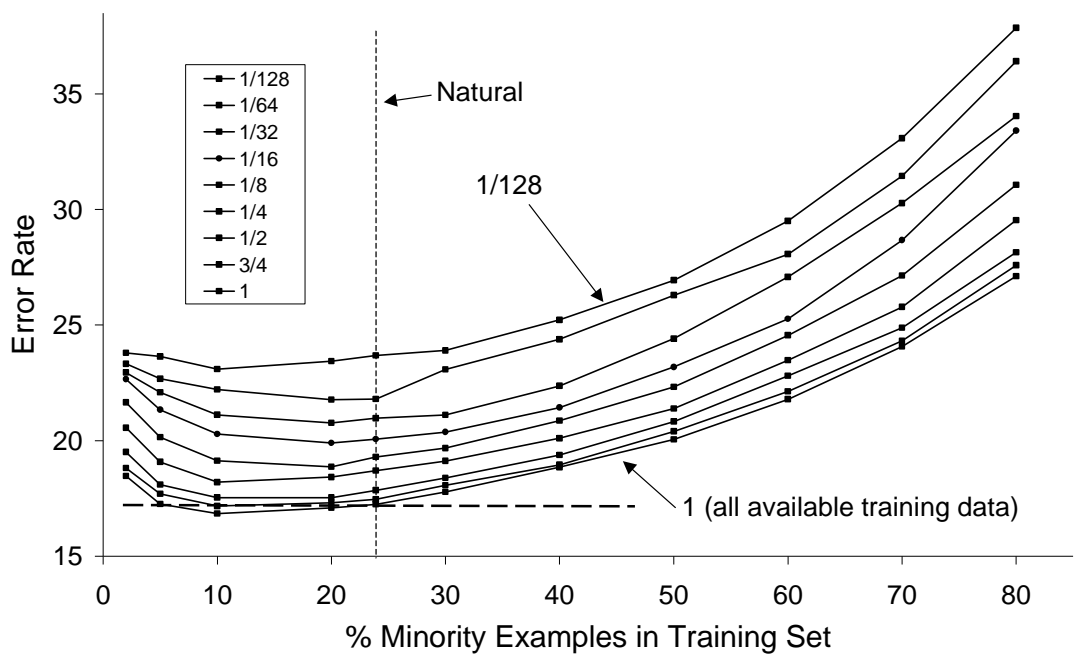

Figure 4: Effect of Class Distribution and Training-set Size on Error Rate (Adult Data Set)

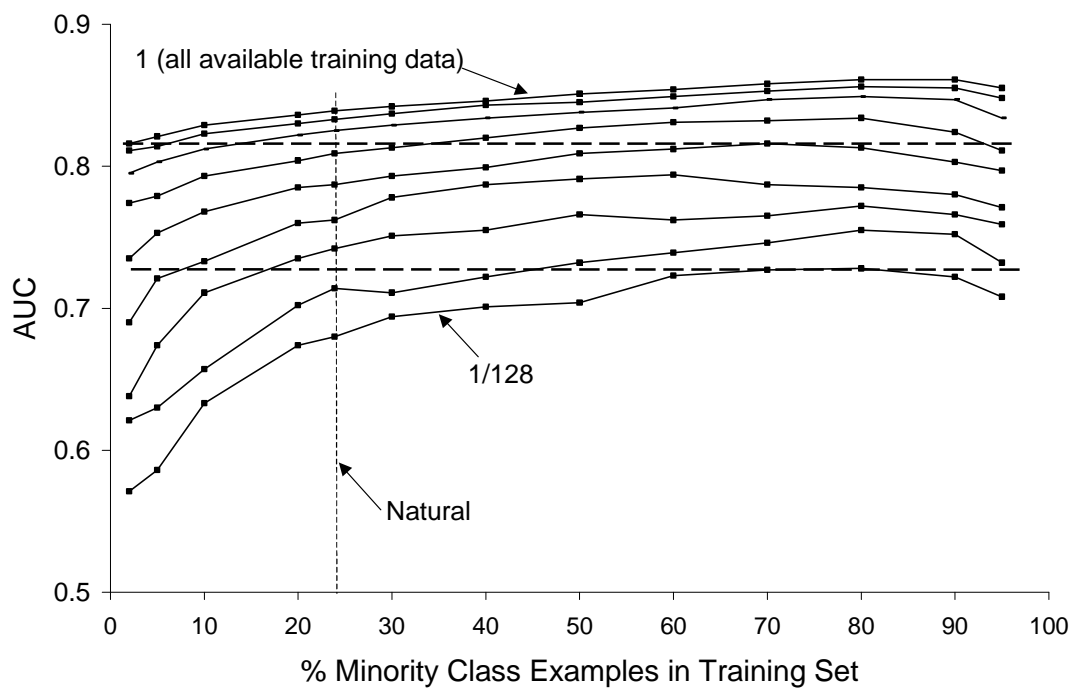

Figure 5: Effect of Class Distribution and Training-set Size on AUC (Adult Data Set)

Figure 4 and Figure 5 show several important things. First, while a change in training-set size shifts the performance curves, the relative rank of each point on each performance curve remains roughly the same. Thus, while the class distribution that yields the best performance occasionally varies with training-set size, these variations are relatively rare and when they occur, they are small. For example, Figure 5 (and the supporting details in Appendix B) indicates that for the adult data set, the class distribution that yields the best AUC typically contains 80% minority-class examples, although there is occasionally a small deviation (with 1/8 the training data 70% minority-class examples does best). This gives support to the notion that there may be a "best" marginal class distribution for a learning task and suggests that a progressive sampling algorithm may be useful in locating the class distribution that yields the best, or nearly best, classifier performance.

The results also indicate that, for any fixed class distribution, increasing the size of the training set always leads to improved classifier performance. Also note that the performance curves tend to "flatten out" as the size of the data set grows, indicating that the choice of class distribution may become less important as the training-set size grows. Nonetheless, even when all of the available training data are used, the choice of class distribution does make a difference. This is significant because if a plateau had been reached (i.e., learning had stopped), then it would be possible to reduce the size of the training set without degrading classifier performance. In that case it would not be necessary to select the class distribution of the training data carefully.

The results in Figure 4 and Figure 5 also show that by carefully selecting the class distribution, one can sometimes achieve improved performance while using fewer training examples. To see this, consider the dashed horizontal line in Figure 4, which intersects the curve associated with ¾ of the training data at its lowest error rate, when the class distribution includes 10% minority-class examples. When this horizontal line is below the curve associated with all available training data, then the training set with ¾ of the data outperforms the full training set. In this case we see that ¾ of the training data with a 10% class distribution outperforms the natural class distribution using all of the available training data. The two horizontal lines in Figure 5 highlight just some of the cases where one can achieve improved AUC using fewer training data (because larger AUC values indicate improved performance, compare the horizontal lines with the curves that lie above them). For example, Figure 5 shows that the training set with a class distribution that contains 80% minority-class examples and $1/128^{th}$ of the total training data outperforms a training set with twice the training data when its class distribution contains less than or equal to 40% minority-class examples (and outperforms a training set with four times the data if its class distribution contains less than or equal to 10% minority-class examples). The results in Appendix B show that all of the trends noted for the adult data set hold for the other data sets and that one can often achieve improved performance using less training data.

## 7.2 A Budget-Sensitive Progressive sampling Algorithm for Selecting Training Data

As discussed above, the size of the training set sometimes must be limited due to costs associated with procuring usable training examples. For simplicity, assume that there is a budget $n$, which permits one to procure exactly $n$ training examples. Further assume that the number of training examples that potentially can be procured is sufficiently large so that a training set of size $n$ can be formed with any desired marginal class distribution. We would like a sampling strategy that selects $x$ minority-class examples and $y$ majority-class examples, where $x + y = n$, such that the resulting class distribution yields the best possible classification performance for a training set of size *n*.

The sampling strategy relies on several assumptions. First, we assume that the cost of executing the learning algorithm is negligible compared to the cost of procuring examples, so that the learning algorithm may be run multiple times. This certainly will be true when training data are costly. We further assume that the cost of procuring examples is the same for each class and hence the budget *n* represents the number of examples that can be procured as well as the total cost. This assumption will hold for many, but not all, domains. For example, for the phone data set described in Section 1 the cost of procuring business and consumer "examples" is equal, while for the telephone fraud domain the cost of procuring fraudulent examples may be substantially higher than the cost of procuring non-fraudulent examples. The algorithm described in this section can be extended to handle non-uniform procurement costs.

The sampling algorithm selects minority-class and majority-class training examples such that the resulting class distribution will yield classifiers that tend to perform well. The algorithm begins with a small amount of training data and progressively adds training examples using a geometric sampling schedule (Provost, Jensen & Oates, 1999). The proportion of minority-class examples and majority-class examples added in each iteration of the algorithm is determined empirically by forming several class distributions from the currently available training data, evaluating the classification performance of the resulting classifiers, and then determining the class distribution that performs best. The algorithm implements a beam-search through the space of possible class distributions, where the beam narrows as the budget is exhausted.

We say that the sampling algorithm is *budget-efficient* if all examples selected during any iteration of the algorithm are used in the final training set, which has the heuristically determined class distribution. The key is to constrain the search through the space of class distributions so that budget-efficiency is either guaranteed, or very likely. As we will show, the algorithm described in this section is guaranteed to be budget-efficient. Note, however, that the class distribution of the final training set, that is heuristically determined, is not guaranteed to be the best class distribution; however, as we will show, it performs well in practice.

The algorithm is outlined in Table 7, using pseudo-code, followed by a line-by-line explanation (a complete example is provided in Appendix C). The algorithm takes three user-specified input parameters: μ, the geometric factor used to determine the rate at which the training-set size grows; *n*, the budget; and *cmin*, the minimum fraction of minority-class examples and majority-class examples that are assumed to appear in the final training set in order for the budget-efficiency guarantee to hold.[5] For the results presented in this section, μ is set to 2, so that the training-set size doubles every iteration of the algorithm, and *cmin* is set to 1/32.

The algorithm begins by initializing the values for the *minority* and *majority* variables, which represent the total number of minority-class examples and majority-class examples requested by the algorithm. Then, in line 2, the number of iterations of the algorithm is determined, such that the initial training-set size, which is subsequently set in line 5, will be at most *cmin • n*. This will allow all possible class distributions to be formed using at most *cmin* minority-class examples and *cmin* majority-class examples. For example, given that μ is 2 and *cmin* is 1/32, in line 2 variable *K* will be set to 5 and in line 5 the initial training-set size will be set to 1/32 *n*.

---

5. Consider the degenerate case where the algorithm determines that the best class distribution contains no minority-class examples or no majority-class examples. If the algorithm begins with even a single example of this class, then it will not be budget-efficient.

1. minority = majority = 0;           # current number minority/majority examples in hand

2.   $K = \left\lceil \log_\mu \left( \frac{1}{c \min} \right) \right\rceil$ ;                           # number of iterations is K+1

3.   for (j = 0; j ≤ K; j = j+1)                    # for each iteration (e.g., j = 0, 1,2,3,4,5)
4.   {
5.       size = $n(1/\mu)^{K-j}$                      # set training-set size for iteration j

6.      if (j = 0)
7.          beam_bottom = 0;   beam_top = 1;        # initialize beam for first iteration
8.      else if (j = K)
9.          beam_bottom = best;  beam_top = best; # last iteration only evaluate previous best
10.     else
11.           beam_radius = $\frac{\min(best, 1 - best)}{\mu + 1 / \mu - 1}$
12.           beam_bottom = best – beam_radius;  beam_top = best + beam_radius;

13.      min_needed = size • beam_top;              # number minority examples needed
14.      maj_needed = size • (1.0 – beam_bottom);  # number majority examples needed

15.      if (min_needed > minority)
16.          request (min_needed - minority) additional minority-class examples;
17.      if (maj_needed > majority)
18.          request (maj_needed - majority) additional majority-class examples;

19.       evaluate(beam_bottom, beam_top, size);  # evaluate distributions in the beam; set best
20.   }

Table 7: The Algorithm for Selecting Training Data

Next, in lines 6-12, the algorithm determines the class distributions to be considered in each iteration by setting the boundaries of the beam. For the first iteration, all class distributions are considered (i.e., the fraction of minority-class examples in the training set may vary between 0 and 1) and for the very last iteration, only the best-performing class distribution from the previous iteration is evaluated. In all other iterations, the beam is centered on the class distribution that performed best in the previous iteration and the radius of the beam is set (in line 11) such that the ratio beam_top/beam_bottom will equal μ. For example, if μ is 2 and *best* is .15, then *beam_radius* is .05 and the beam will span from .10 to .20—which differ by a factor of 2 (i.e., μ).

In lines 13 and 14 the algorithm computes the number of minority-class examples and majority-class examples needed to form the class distributions that fall within the beam. These values are determined from the class distributions at the boundaries of the beam. In lines 15-18 additional examples are requested, if required. In line 19 an evaluation procedure is called to form the class distributions within the beam and then to induce and to evaluate the classifiers. At a minimum this procedure will evaluate the class distributions at the endpoints and at the midpoint of the beam; however, this procedure may be implemented to evaluate additional class distributions within the beam. The procedure will set the variable *best* to the class distribution that performs best. If the best performance is achieved by several class distributions, then a resolution procedure is needed. For example, the class distribution for which the surrounding class distributions perform best may be chosen; if this still does not yield a unique value, then the best-performing class distribution closest to the center of the beam may be chosen. In any event, for the last iteration, only one class distribution is evaluated—the previous best. To ensure budget-efficiency, only one class distribution can be evaluated in the final iteration.

This algorithm is guaranteed to request only examples that are subsequently used in the final training set, which will have the heuristically determined class distribution. This guarantee can be verified inductively. First, the base case. The calculation for $K$ in line 2 ensures that the initial training set will contain $cmin \cdot n$ training examples. Since we assume that the final training set will have at least $cmin$ minority-class examples and $cmin$ majority-class examples, all examples used to form the initial training set are guaranteed to be included in the final training set. Note that $cmin$ may be set arbitrarily small—the smaller $cmin$ the larger $K$ and the smaller the size of the initial training set.

The inductive step is based on the observation that because the radius of the beam in line 11 is set so that the beam spans at most a factor of $\mu$, all examples requested in each iteration are guaranteed to be used in the final training set. To see this, we will work backward from the final iteration, rather than working forward as is the case in most inductive proofs. Assume that the result of the algorithm is that the fraction of minority-class examples in the final training set is $p$, so that there are $p \cdot n$ minority-class examples in the final training set. This means that $p$ was the best distribution from the previous iteration. Since $p$ must fall somewhere within the beam for the previous iteration and the beam must span a factor $\mu$, we can say the following: the fraction of minority-class examples in the previous iteration could range from $p/\mu$ (if $p$ was at the top of the previous beam) to $\mu \cdot p$ (if $p$ was at the bottom of the previous beam). Since the previous iteration contains $n/\mu$ examples, due to the geometric sampling scheme, then the previous iteration has at most $(\mu \cdot p) \cdot n/\mu$, or $p \cdot n$, minority-class examples. Thus, in all possible cases all minority-class examples from the previous iteration can be used in the final interaction. This argument applies similarly to the majority-class examples and can be extended backwards to previous iterations.[6] Thus, because of the bound on the initial training-set size and the restriction on the width of the beam not to exceed the geometric factor $\mu$, the algorithm guarantees that all examples requested during the execution of the algorithm will be used in the final training set.

A complete, detailed, iteration-by-iteration example describing the sampling algorithm as it is applied to the phone data set is provided in Appendix C, Table C1. In that example error rate is used to evaluate classifier performance. The description specifies the class distributions that are evaluated during the execution of the algorithm. This "trajectory" is graphically depicted in Figure 6, narrowing in on the final class distribution. At each iteration, the algorithm considers the "beam" of class distributions bounded by the two curves.

---

6. The only exception is for the first iteration of the algorithm, since in this situation the beam is unconditionally set to span all class distributions. This is the reason why the *cmin* value is required to provide the efficiency guarantee.

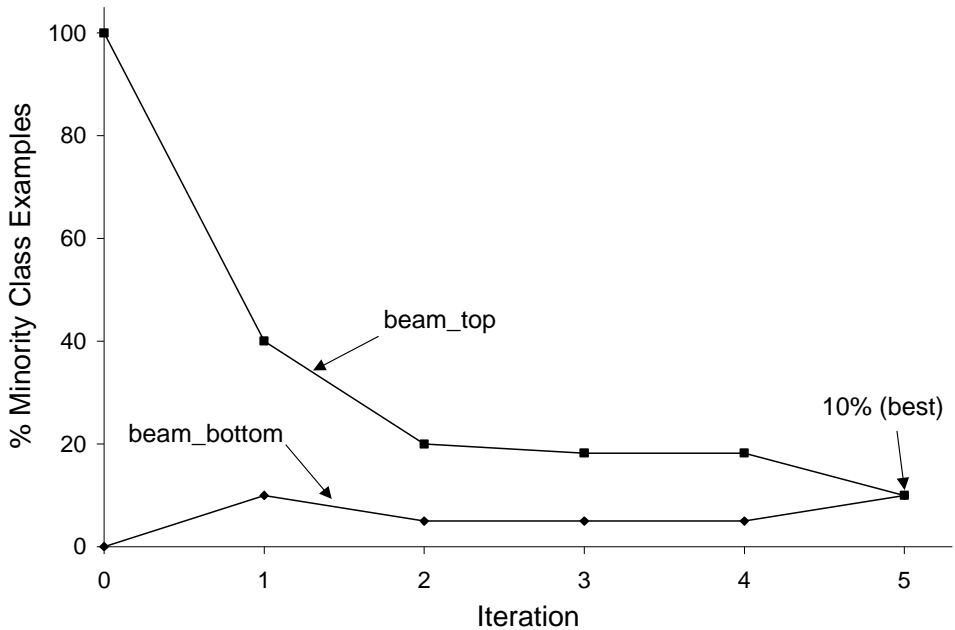

Figure 6: The Trajectory of the Algorithm through the Space of Class Distributions.

## 7.3 Results for the Sampling Algorithm

The budget-sensitive progressive sampling algorithm was applied to the phone, adult, covertype, kr-vs-kp, weather, letter-a and blackjack data sets using both error rate and AUC to measure classifier performance. However, the method for setting the beam (described in lines 6-12 in Table 7) was modified so that the results from the experiments described in Section 7.1 (and detailed in Appendix B), which evaluate only the 13 listed class distributions, could be used. Specifically, at each iteration the low end (high end) of the beam is set to the class distribution specified in Appendix B that is just below (above) the best performing class distribution from the prior iteration. For example, if in one iteration the best performing class distribution contains 30% minority-class examples, then in the next iteration the bottom of the beam is set to include 20% minority-class examples and the top of the beam to include 40% minority-class examples (of the 13 sampled class distributions, these are the closest to the 30% class distribution). Although this will sometimes allow the beam to span a range greater than $\mu$ (2), in practice this does not result in a problem—for the seven data sets all examples requested by the algorithm are included in the final training set. In addition, a slight improvement was made to the algorithm. Specifically, for any iteration, if the number of examples already in hand (procured in previous iterations) is sufficient to evaluate additional class distributions, then the beam is widened to include these additional class distributions (this can happen because during the first iteration the beam is set very wide).

The performance of this sampling algorithm is summarized in Table 8, along with the performance of two other strategies for selecting the class distribution of the training data. The first of the two additional strategies, the "Pick Natural/Balanced Strategy," is based on the guidelines suggested by the empirical results from Section 6. This strategy selects the natural distribution when error rate is the performance metric and the balanced class distribution when AUC is the performance metric. The "Pick Best" strategy selects the class distribution that performs best over the 13 evaluated class distributions (see Tables 5 and 6). Given that we only consider the 13 class distributions, this strategy will always yield the best results but, as we shall see, is more costly than the other strategies. The value representing the best, budget-efficient performance (lowest error rate, highest AUC) is underlined for each data set. A detailed iteration-by-iteration description of the algorithm, for all seven data sets, is provided in Appendix C, Table C3.

Table 8 also specifies the cost for each strategy, based on the number of training examples requested by the algorithm. This cost is expressed with respect to the budget $n$ (each strategy yields a final training set with $n$ examples). The "Pick Natural/Balanced" strategy always requires exactly $n$ examples to be selected and therefore has a cost of $n$ and is budget-efficient. The "Pick Best" strategy has a total cost of $1.93n$ and hence is not budget-efficient (because it evaluates class distributions with between 2% and 95% minority-class examples, it requires $.95n$ minority-class examples and $.98n$ majority-class examples). The cost of the sampling algorithm depends on the performance of the induced classifiers: with the changes to the algorithm described in this section, it is no longer guaranteed to be budget-efficient. Nonetheless, in all cases—for both error rate and AUC—the sampling algorithm has a cost of exactly $n$ and hence turns out to be budget-efficient.

| Data Set | Sampling Algorithm | | | Pick Natural/Balanced | | | Pick Best | | |
|---|---|---|---|---|---|---|---|---|---|
| | ER | AUC | Cost | ER | AUC | Cost | ER | AUC | Cost |
| phone | 12.3% | .851 | $n$ | 12.6% | .849 | $n$ | 12.3% | .853 | $1.93n$ |
| adult | 17.1% | .861 | $n$ | 17.3% | .851 | $n$ | 16.9% | .861 | $1.93n$ |
| covertype | 5.0% | .984 | $n$ | 5.0% | .980 | $n$ | 5.0% | .984 | $1.93n$ |
| kr-vs-kp | 1.2% | .998 | $n$ | 1.4% | .998 | $n$ | 1.2% | .998 | $1.93n$ |
| weather | 33.1% | .740 | $n$ | 33.7% | .736 | $n$ | 33.1% | .740 | $1.93n$ |
| letter-a | 2.8% | .954 | $n$ | 2.8% | .937 | $n$ | 2.6% | .954 | $1.93n$ |
| blackjack | 28.4% | .715 | $n$ | 28.4% | .713 | $n$ | 28.4% | .715 | $1.93n$ |

Table 8: Comparative Performance of the Sampling Algorithm

The results in Table 8 show that by using the budget-sensitive progressive sampling algorithm to choose the training data it is possible to achieve results that are as good as or better than the strategy of always using the natural distribution for error rate and the balanced distribution for AUC—without requiring that any extra examples be procured. In particular, when comparing these two strategies, the progressive sampling strategy has a win-tie-loss record of 10-4-0. While in some cases these wins do not lead to large improvements in performance, in some cases they do (e.g., for the kr-vs-kp data set the sampling strategy yields a relative reduction in error rate of 17%). The results in Table 8 also show that the sampling algorithm performs nearly as well as the "Pick Best" strategy (it performs as well in 11 of 14 cases), which is almost twice as costly. Because the progressive sampling strategy performs nearly as well as the "Pick Best" strategy, we conclude that when the progressive sampling strategy does not substantially outperform the "Pick Natural/Balanced" strategy, it is not because the sampling strategy cannot identify a good (i.e., near-optimal) class distribution for learning, but rather that the optimal class distribution happens to be near the natural (balanced) distribution for error rate (AUC). Note that there are some data sets (optdigits, contraceptive, solar-flare, car) for which this is not the case and hence the "Pick Natural/Balanced" strategy will perform poorly. Unfortunately, because these data sets would yield relatively small training sets, the progressive sampling algorithm could not be run on them.

In summary, the sampling algorithm introduced in this section leads to near-optimal results—results that outperform the straw-man strategy of using the natural distribution to minimize error rate and the balanced distribution to maximize AUC. Based on these results, the budget-sensitive progressive sampling algorithm is attractive—it incurs the minimum possible cost in terms of procuring examples while permitting the class distribution for training to be selected using some intelligence.

## 8. Related Work

Several researchers have considered the question of what class distribution to use for a fixed training-set size, and/or, more generally, how class distribution affects classifier performance. Both Catlett (1991) and Chan & Stolfo (1998) study the relationship between (marginal) training class distribution and classifier performance when the training-set size is held fixed, but focus most of their attention on other issues. These studies also analyze only a few data sets, which makes it impossible to draw general conclusions about the relationship between class distribution and classifier performance. Nonetheless, based on the results for three data sets, Chan & Stolfo (1998) show that when accuracy is the performance metric, a training set that uses the natural class distribution yields the best results. These results agree partially with our results—although we show that the natural distribution does not always maximize accuracy, we show that the optimal distribution generally is close to the natural distribution. Chan & Stolfo also show that when actual costs are factored in (i.e., the cost of a false positive is not the same as a false negative), the natural distribution does not perform best; rather a training distribution closer to a balanced distribution performs best. They also observe, as we did, that by increasing the percentage of minority-class examples in the training set, the induced classifier performs better at classifying minority examples. It is important to note, however, that neither Chan & Stolfo nor Catlett adjusted the induced classifiers to compensate for changes made to the class distribution of the training set. This means that their results are biased in favor of the natural distribution (when measuring classification accuracy) and that they could improve the classification performance of minority class examples simply by changing (implicitly) the decision threshold. As the results in Appendix A show, compensating for the changed class distribution can affect the performance of a classifier significantly.

Several researchers have looked at the general question of how to reduce the need for labeled training data by selecting the data intelligently, but without explicitly considering the class distribution. For example, Cohn et al. (1994) and Lewis and Catlett (1994) use "active learning" to add examples to the training set for which the classifier is least certain about the classification. Saar-Tsechansky and Provost (2001, 2003) provide an overview of such methods and also extend them to cover AUC and other non-accuracy based performance metrics. The setting where these methods are applicable is different from the setting we consider. In particular, these methods assume either that arbitrary examples can be labeled or that the descriptions of a pool of unlabeled examples are available and the critical cost is associated with labeling them (so the algorithms select the examples intelligently rather than randomly). In our typical setting, the cost is in procuring the descriptions of the examples—the labels are known beforehand.

There also has been some prior work on progressive sampling strategies. John and Langley (1996) show how one can use the extrapolation of learning curves to determine when classifier performance using a subset of available training data comes close to the performance that would be achieved by using the full data set. Provost et al. (1999) suggest using a geometric sampling schedule and show that it is often more efficient than using all of the available training data. The techniques described by John and Langley (1996) and Provost et al. (1999) do not change the distribution of examples in the training set, but rather rely on taking random samples from the available training data. Our progressive sampling routine extends these methods by stratifying the sampling by class, and using the information acquired during the process to select a good final class distribution.

There is a considerable amount of research on how to build "good" classifiers when the class distribution of the data is highly unbalanced and it is costly to misclassify minority-class examples (Japkowicz et al., 2000). This research is related to our work because a frequent approach for learning from highly skewed data sets is to modify the class distribution of the training set. Under these conditions, classifiers that optimize for accuracy are especially inappropriate because they tend to generate trivial models that almost always predict the majority class. A common approach for dealing with highly unbalanced data sets is to reduce the amount of class imbalance in the training set. This tends to produce classifiers that perform better on the minority class than if the original distribution were used. Note that in this situation the training-set size is not fixed and the motivation for changing the distribution is simply to produce a "better" classifier—not to reduce, or minimize, the training-set size.

The two basic methods for reducing class imbalance in training data are under-sampling and over-sampling. Under-sampling eliminates examples in the majority class while over-sampling replicates examples in the minority class (Breiman, et al., 1984; Kubat & Matwin, 1997; Japkowicz & Stephen, 2001). Neither approach consistently outperforms the other nor does any specific under-sampling or over-sampling rate consistently yield the best results. Estabrooks and Japkowicz (2001) address this issue by showing that a mixture-of-experts approach, which combines classifiers built using under-sampling and over-sampling methods with various sampling rates, can produce consistently good results.

Both under-sampling and over-sampling have known drawbacks. Under-sampling throws out potentially useful data while over-sampling *increases the size of the training set* and hence the time to build a classifier. Furthermore, since most over-sampling methods make exact copies of minority class examples, overfitting is likely to occur—classification rules may be induced to cover a single replicated example.[7] Recent research has focused on improving these basic methods. Kubat and Matwin (1997) employ an under-sampling strategy that intelligently removes majority examples by removing only those majority examples that are "redundant" or that "border" the minority examples—figuring they may be the result of noise. Chawla et al. (2000) combine under-sampling and over-sampling methods, and, to avoid the overfitting problem, form new minority class examples by interpolating between minority-class examples that lie close together. Chan and Stolfo (1998) take a somewhat different, and innovative, approach. They first run preliminary experiments to determine the best class distribution for learning and then generate multiple training sets with this class distribution. This is typically accomplished by including all minority-class examples and some of the majority-class examples in each training set. They then apply a learning algorithm to each training set and then combine the generated classifiers to form a composite learner. This method ensures that all available training data are used, since each majority-class example will be found in at least one of the training sets.

---

7. This is especially true for methods such as C4.5, which stops splitting based on counting examples at the leaves of the tree.

The research in this article could properly be viewed as research into under-sampling and its effect on classifier performance. However, given this perspective, our research performs under-sampling in order to reduce the training-set size, whereas in the research relating to skewed data sets the primary motivation is to improve classifier performance. For example, Kubat and Matwin (1997) motivate the use of under-sampling to handle skewed data sets by saying that "adding examples of the majority class to the training set can have a detrimental effect on the learner's behavior: noisy or otherwise unreliable examples from the majority class can overwhelm the minority class" (p. 179). A consequence of these different motivations is that in our experiments we under-sample the minority and/or majority classes, while in the research concerned with learning from skewed distributions it is only the majority class that is under-sampled.

The use of under-sampling for reducing the training-set size (and thereby reducing cost) may be the more practically useful perspective. Reducing the class imbalance in the training set effectively causes the learner to impose a greater cost for misclassifying minority-class examples (Breiman et al., 1984). Thus, when the cost of acquiring and learning from the data is not an issue, cost-sensitive or probabilistic learning methods are a more direct and arguably more appropriate way of dealing with class imbalance, because they do not have the problems, noted earlier, that are associated with under-sampling and over-sampling. Such approaches have been shown to outperform under-sampling and over-sampling (Japkowicz & Stephen, 2002). To quote Drummond and Holte (2000) "all of the data available can be used to produce the tree, thus throwing away no information, and learning speed is not degraded due to duplicate instances" (p. 239).

## 9. Limitations and Future Research

One limitation with the research described in this article is that because all results are based on the use of a decision-tree learner, our conclusions may hold only for this class of learners. However, there are reasons to believe that our conclusions will hold for other learners as well. Namely, since the role that class distribution plays in learning—and the reasons, discussed in Section 5.2, for why a classifier will perform worse on the minority class—are not specific to decision-tree learners, one would expect other learners to behave similarly. One class of learners that may especially warrant further attention, however, are those learners that do not form disjunctive concepts. These learners will not suffer in the same way from the "problem of small disjuncts," which our results indicate is partially responsible for minority-class predictions having a higher error rate than majority-class predictions.[8] Thus, it would be informative to extend this study to include other classes of learners, to determine which results indeed generalize.

The program for inducing decision trees used throughout this article, C4.5, only considers the class distribution of the training data when generating the decision tree. The differences between the class distribution of the training data and the test data are accounted for in a post-processing step by re-computing the probability estimates at the leaves and using these estimates to re-label the tree. If the induction program had knowledge of the target (i.e., test) distribution during the tree-building process, then a different decision tree might be constructed. However, research indicates that this is not a serious limitation. In particular, Drummond and Holte (2000) showed that there are splitting criteria that are completely insensitive to the class distribution and that these splitting criteria perform as well or better than methods that factor in the class distribution. They further showed that C4.5's splitting criterion is relatively insensitive to the class distribution—and therefore to changes in class distribution.

---

8. However, many learners do form disjunctive concepts or something quite close. For example, Van den Bosch et al. (1997) showed that instance-based learners can be viewed as forming disjunctive concepts.

We employed C4.5 without pruning in our study because pruning is sensitive to class distribution and C4.5's pruning strategy does not take the changes made to the class distribution of the training data into account. To justify this choice we showed that C4.5 without pruning performs competitively with C4.5 with pruning (Sections 6.2 and 6.3). Moreover, other research (Bradford et al., 1998) indicates that classifier performance does not generally improve when pruning takes class distribution and costs into account. Nevertheless it would be worthwhile to see just how a "cost/distribution-sensitive" pruning strategy would affect our results. We know of no published pruning method that attempts to maximize AUC.

In this article we introduced a budget-sensitive algorithm for selecting training data when it is costly to obtain usable training examples. It would be interesting to consider the case where it is more costly to procure examples belonging to one class than to another.

## 10. Conclusion

In this article we analyze, for a fixed training-set size, the relationship between the class distribution of training data and classifier performance with respect to accuracy and AUC. This analysis is useful for applications where data procurement is costly and data can be procured independently by class, or where the costs associated with learning from the training data are sufficient to require that the size of the training set be reduced. Our results indicate that when accuracy is the performance measure, the best class distribution for learning tends to be near the natural class distribution, and when AUC is the performance measure, the best class distribution for learning tends to be near the balanced class distribution. These general guidelines are just that—guidelines—and for a particular data set a different class distribution may lead to substantial improvements in classifier performance. Nonetheless, if no additional information is provided and a class distribution must be chosen without any experimentation, our results show that for accuracy and for AUC maximization, the natural distribution and a balanced distribution (respectively) are reasonable default training distributions.

If it is possible to interleave data procurement and learning, we show that a budget-sensitive progressive sampling strategy can improve upon the default strategy of using the natural distribution to maximize accuracy and a balanced distribution to maximize the area under the ROC curve—in our experiments the budget-sensitive sampling strategy never did worse. Furthermore, in our experiments the sampling strategy performs nearly as well as the strategy that evaluates many different class distributions and chooses the best-performing one (which is optimal in terms of classification performance but inefficient in terms of the number of examples required).

The results presented in this article also indicate that for many data sets the class distribution that yields the best-performing classifiers remains relatively constant for different training-set sizes, supporting the notion that there often is a "best" marginal class distribution. These results further show that as the amount of training data increases the differences in performance for different class distributions lessen (for both error rate and AUC), indicating that as more data becomes available, the choice of marginal class distribution becomes less and less important—especially in the neighborhood of the optimal distribution.

This article also provides a more comprehensive understanding of how class distribution affects learning and suggests answers to some fundamental questions, such as why classifiers almost always perform worse at classifying minority-class examples. A method for adjusting a classifier to compensate for changes made to the class distribution of the training set is described and this adjustment is shown to substantially improve classifier accuracy (see Appendix A). We consider this to be particularly significant because previous research on the effect of class distribution on learning has not employed this, or any other, adjustment (Catlett, 1991; Chan & Stolfo, 1998; Japkowicz & Stephen, 2002).

Practitioners often make changes to the class distribution of training data, especially when the classes are highly unbalanced. These changes are seldom done in a principled manner and the reasons for changing the distribution—and the consequences—are often not fully understood. We hope this article helps researchers and practitioners better understand the relationship between class distribution and classifier performance and permits them to learn more effectively when there is a need to limit the amount of training data.

## Acknowledgments

We would like to thank Haym Hirsh for the comments and feedback he provided throughout this research. We would also like to thank the anonymous reviewers for helpful comments, and IBM for a Faculty Partnership Award.

## Appendix A: Impact of Class Distribution Correction on Classifier Performance

Table A1 compares the performance of the decision trees labeled using the uncorrected frequency-based estimate (FB) with those labeled using the corrected frequency-based estimate (CT-FB).

| Dataset | Error Rate | | % Rel. | % Labels | % Errors from Min. | |
|---|---|---|---|---|---|---|
| | FB | CT-FB | Improv. | Changed | FB | CT-FB |
| letter-a | 9.79 | 5.38 | 45.0 | 39.0 | 2.7 | 7.2 |
| pendigits | 4.09 | 4.02 | 1.7 | 3.2 | 5.6 | 7.8 |
| abalone | 30.45 | 22.97 | 24.6 | 5.6 | 8.5 | 19.1 |
| sick-euthyroid | 9.82 | 6.85 | 30.2 | 6.7 | 8.8 | 14.6 |
| connect-4 | 30.21 | 27.57 | 8.7 | 14.7 | 8.5 | 10.4 |
| optdigits | 6.17 | 3.41 | 44.7 | 42.5 | 6.0 | 21.2 |
| covertype | 6.62 | 6.46 | 2.4 | 2.4 | 7.0 | 8.5 |
| solar-flare | 36.20 | 29.12 | 19.6 | 20.4 | 19.3 | 30.7 |
| phone | 17.85 | 14.81 | 17.0 | 3.2 | 25.2 | 44.4 |
| letter-vowel | 18.89 | 14.16 | 25.0 | 44.1 | 15.9 | 30.2 |
| contraceptive | 40.77 | 39.65 | 2.7 | 11.1 | 20.6 | 27.6 |
| adult | 22.69 | 20.05 | 11.6 | 30.7 | 19.6 | 36.8 |
| splice-junction | 9.02 | 8.74 | 3.1 | 14.1 | 20.1 | 28.4 |
| network2 | 30.80 | 29.96 | 2.7 | 1.2 | 32.9 | 40.1 |
| yeast | 34.01 | 28.80 | 15.3 | 4.6 | 29.4 | 47.0 |
| network1 | 31.99 | 30.99 | 3.1 | 1.3 | 32.9 | 38.2 |
| car | 8.26 | 7.92 | 4.1 | 5.3 | 25.9 | 33.8 |
| german | 38.37 | 37.09 | 3.3 | 16.1 | 30.8 | 35.8 |
| breast-wisc | 6.76 | 6.74 | 0.3 | 0.4 | 38.5 | 38.7 |
| blackjack | 33.02 | 28.71 | 13.1 | 17.1 | 42.9 | 76.2 |
| weather | 34.62 | 34.61 | 0.0 | 0.0 | 40.5 | 40.5 |
| bands | 32.68 | 32.68 | 0.0 | 0.6 | 90.2 | 90.2 |
| market1 | 25.77 | 25.77 | 0.0 | 23.9 | 46.0 | 48.6 |
| crx | 20.84 | 21.48 | -3.1 | 17.2 | 46.2 | 51.4 |
| kr-vs-kp | 1.22 | 1.22 | 0.0 | 0.2 | 58.5 | 58.5 |
| move | 28.24 | 28.24 | 0.0 | 20.8 | 52.6 | 60.7 |
| Average | 21.89 | 19.90 | 10.6 | 13.3 | 28.3 | 36.4 |

Table A1: Impact of the Probability Estimates on Error Rate

The results in the main body of the article were all based on the use of the corrected frequency-based estimate to label the leaves of the induced decision trees, so that the decision trees were not improperly biased by the changes made to the class distribution of the training set. Thus, the comparison in Table A1 evaluates the significance of correcting for changes to the class distribution of the training data. This comparison is based on the situation where the class distribution of the training set is altered to contain an equal number of minority-class and majority-class examples (the test set will still contain the natural class distribution). The results are based on 30 runs and the data sets are listed in order of decreasing class imbalance.

The error rate for the estimates is displayed in the second and third columns in Table A1, and, for each data set, the lowest error rate is underlined. The fourth column specifies the relative improvement that results from using the corrected frequency-based estimate. The fifth column specifies the percentage of the leaves in the decision tree that are assigned a different class label when the corrected estimate is used. The last two columns specify, for each estimate, the percentage of the total errors that are contributed by the minority-class test examples.

Table A1 shows that by employing the corrected frequency-based estimate instead of the uncorrected frequency-based estimate, there is, on average, a relative 10.6% reduction in error rate. Furthermore, in only one case does the uncorrected frequency-based estimate outperform the corrected frequency-based estimate. The correction tends to yield a larger reduction for the most highly unbalanced data sets—in which cases it plays a larger role. If we restrict ourselves to the first 13 data sets listed in Table 2, for which the minority class makes up less than 25% of the examples, then the relative improvement over these data sets is 18.2%. Note that because in this scenario the minority class is over-sampled in the training set, the corrected frequency-based estimate can only cause minority-labeled leaves to be labeled with the majority-class. Consequently, as the last column in the table demonstrates, the corrected version of the estimate will cause more of the errors to come from the minority-class test examples.

## Appendix B: The Effect of Training-Set Size and Class Distribution on Learning

Experiments were run to establish the joint impact that class distribution and training-set size have on classifier performance. Classifier performance is reported for the same thirteen class distributions that were analyzed in Section 6 and for nine different training set sizes. The nine training set sizes are generated by omitting a portion of the available training data (recall that, as described in Section 4.1, the amount of available training data equals ¾ of the number of minority-class examples). For these experiments the training set sizes are varied so as to contain the following fractions of the total available training data: 1/128, 1/64, 1/32, 1/16, 1/8, 1/4, 1/2, 3/4 and 1. In order to ensure that the training sets contain a sufficient number of training examples to provide meaningful results, the original data set must be relatively large and/or contain a high proportion of minority-class examples. For this reason, only the following seven data sets were selected for analysis: phone, adult, covertype, kr-vs-kp, weather, letter-a and blackjack. Because the last four data sets in this list yield a smaller number of training examples than the first three, for these data sets the two smallest training-set sizes (1/128 and 1/64) are not evaluated. The experimental results are summarized in Tables B1a and B1b. An asterisk is used to denote the natural class distribution for each data set and, for each training-set size, the class distribution that yields the best performance is displayed in bold and is underlined.

**PHONE**

| Size | Metric | 2 | 5 | 10 | 18.2* | 20 | 30 | 40 | 50 | 60 | 70 | 80 | 90 | 95 |
|---|---|---|---|---|---|---|---|---|---|---|---|---|---|---|
| 1/128 | AUC | .641 | .737 | .784 | **.793** | .792 | .791 | .791 | .789 | .788 | .786 | .785 | .774 | .731 |
| 1/64 | | .707 | .777 | .784 | **.803** | **.803** | **.803** | .801 | .802 | .801 | .798 | .799 | .788 | .744 |
| 1/32 | | .762 | .794 | .809 | **.812** | **.812** | .811 | .811 | .811 | .810 | **.812** | .811 | .805 | .778 |
| 1/16 | | .784 | .813 | .816 | .823 | .823 | **.824** | .818 | .821 | .822 | .821 | .822 | .817 | .805 |
| 1/8 | | .801 | .823 | .828 | .830 | .830 | .830 | .830 | .829 | .830 | .829 | **.831** | .828 | .818 |
| 1/4 | | .819 | .835 | .837 | .839 | .839 | .837 | .836 | .836 | .836 | .836 | **.838** | .836 | .832 |
| 1/2 | | .832 | .843 | **.846** | **.846** | .845 | .845 | .843 | .843 | .843 | .843 | .844 | **.846** | .844 |
| 3/4 | | .838 | .847 | .849 | .849 | .848 | .846 | .847 | .846 | .847 | .848 | .848 | **.851** | .848 |
| 1 | | .843 | .850 | .852 | .851 | .851 | .850 | .850 | .849 | .848 | .848 | .850 | **.853** | .850 |
| 1/128 | Error Rate | 17.47 | 16.42 | **15.71** | 16.10 | 16.25 | 17.52 | 18.81 | 21.21 | 22.87 | 26.40 | 30.43 | 33.26 | 37.27 |
| 1/64 | | 17.01 | 15.75 | 15.21 | **15.12** | 15.20 | 16.39 | 17.59 | 19.60 | 22.11 | 24.80 | 27.34 | 30.21 | 26.86 |
| 1/32 | | 16.22 | 15.02 | 14.52 | **14.50** | 14.75 | 15.41 | 16.81 | 18.12 | 20.02 | 21.77 | 24.86 | 25.31 | 28.74 |
| 1/16 | | 15.78 | 14.59 | **14.01** | 14.02 | 14.18 | 14.70 | 16.09 | 17.50 | 18.68 | 20.70 | 22.46 | 24.15 | 24.52 |
| 1/8 | | 15.17 | 14.08 | **13.46** | 13.61 | 13.71 | 14.27 | 15.30 | 16.51 | 17.66 | 19.66 | 21.26 | 23.23 | 23.33 |
| 1/4 | | 14.44 | 13.55 | **13.12** | 13.23 | 13.27 | 13.85 | 14.78 | 15.85 | 17.09 | 18.94 | 20.43 | 22.28 | 22.90 |
| 1/2 | | 13.84 | 13.18 | **12.81** | 12.83 | 12.95 | 13.47 | 14.38 | 15.30 | 16.43 | 17.88 | 19.57 | 21.68 | 21.68 |
| 3/4 | | 13.75 | 13.03 | **12.60** | 12.70 | 12.74 | 13.35 | 14.12 | 15.01 | 16.17 | 17.33 | 18.82 | 20.43 | 21.24 |
| 1 | | 13.45 | 12.87 | **12.32** | 12.62 | 12.68 | 13.25 | 13.94 | 14.81 | 15.97 | 17.32 | 18.73 | 20.24 | 21.07 |

**ADULT**

| Size | Metric | 2 | 5 | 10 | 20 | 23.9* | 30 | 40 | 50 | 60 | 70 | 80 | 90 | 95 |
|---|---|---|---|---|---|---|---|---|---|---|---|---|---|---|
| 1/128 | AUC | .571 | .586 | .633 | .674 | .680 | .694 | .701 | .704 | .723 | .727 | **.728** | .722 | .708 |
| 1/64 | | .621 | .630 | .657 | .702 | .714 | .711 | .722 | .732 | .739 | .746 | **.755** | .752 | .732 |
| 1/32 | | .638 | .674 | .711 | .735 | .742 | .751 | .755 | .766 | .762 | .765 | **.772** | .766 | .759 |
| 1/16 | | .690 | .721 | .733 | .760 | .762 | .778 | .787 | .791 | **.794** | .787 | .785 | .780 | .771 |
| 1/8 | | .735 | .753 | .768 | .785 | .787 | .793 | .799 | .809 | .812 | **.816** | .813 | .803 | .797 |
| 1/4 | | .774 | .779 | .793 | .804 | .809 | .813 | .820 | .827 | .831 | .832 | **.834** | .824 | .811 |
| 1/2 | | .795 | .803 | .812 | .822 | .825 | .829 | .834 | .838 | .841 | .847 | **.849** | .847 | .834 |
| 3/4 | | .811 | .814 | .823 | .830 | .833 | .837 | .843 | .845 | .849 | .853 | **.856** | .855 | .848 |
| 1 | | .816 | .821 | .829 | .836 | .839 | .842 | .846 | .851 | .854 | .858 | **.861** | **.861** | .855 |
| 1/128 | Error Rate | 23.80 | 23.64 | 23.34 | **23.10** | 23.44 | 23.68 | 25.22 | 26.94 | 29.50 | 33.08 | 37.85 | 46.13 | 48.34 |
| 1/64 | | 23.32 | 22.68 | 22.21 | **21.77** | 21.80 | 23.08 | 24.38 | 26.29 | 28.07 | 31.45 | 36.41 | 43.64 | 47.52 |
| 1/32 | | 22.95 | 22.09 | 21.12 | **20.77** | 20.97 | 21.11 | 22.37 | 24.41 | 27.08 | 30.27 | 34.04 | 42.40 | 47.20 |
| 1/16 | | 22.66 | 21.34 | 20.29 | **19.90** | 20.07 | 20.37 | 21.43 | 23.18 | 25.27 | 28.67 | 33.41 | 40.65 | 46.68 |
| 1/8 | | 21.65 | 20.15 | 19.13 | **18.87** | 19.30 | 19.67 | 20.86 | 22.33 | 24.56 | 27.14 | 31.06 | 38.35 | 45.83 |
| 1/4 | | 20.56 | 19.08 | **18.20** | 18.42 | 18.70 | 19.12 | 20.10 | 21.39 | 23.48 | 25.78 | 29.54 | 36.17 | 43.93 |
| 1/2 | | 19.51 | 18.10 | **17.54** | **17.54** | 17.85 | 18.39 | 19.38 | 20.83 | 22.81 | 24.88 | 28.15 | 34.71 | 41.24 |
| 3/4 | | 18.82 | 17.70 | **17.17** | 17.32 | 17.46 | 18.07 | 18.96 | 20.40 | 22.13 | 24.32 | 27.59 | 33.92 | 40.47 |
| 1 | | 18.47 | 17.26 | **16.85** | 17.09 | 17.25 | 17.78 | 18.85 | 20.05 | 21.79 | 24.08 | 27.11 | 33.00 | 39.75 |

**COVERTYPE**

| Size | Metric | 2 | 5 | 10 | 14.8* | 20 | 30 | 40 | 50 | 60 | 70 | 80 | 90 | 95 |
|---|---|---|---|---|---|---|---|---|---|---|---|---|---|---|
| 1/128 | AUC | .767 | .852 | .898 | .909 | **.916** | .913 | **.916** | **.916** | .909 | .901 | .882 | .854 | .817 |
| 1/64 | | .836 | .900 | .924 | .932 | **.937** | .935 | .936 | .932 | .928 | .922 | .913 | .885 | .851 |
| 1/32 | | .886 | .925 | .942 | .947 | **.950** | .947 | .948 | .948 | .944 | .939 | .930 | .908 | .876 |
| 1/16 | | .920 | .944 | .953 | .957 | **.959** | **.959** | **.959** | .957 | .955 | .951 | .945 | .929 | .906 |
| 1/8 | | .941 | .955 | .963 | .965 | .967 | .968 | **.969** | .968 | .967 | .963 | .957 | .948 | .929 |
| 1/4 | | .953 | .965 | .970 | .973 | .975 | **.976** | .975 | .973 | .972 | .970 | .965 | .956 | .943 |
| 1/2 | | .963 | .972 | .979 | **.981** | **.981** | .980 | .978 | .977 | .975 | .972 | .970 | .961 | .953 |
| 3/4 | | .968 | .976 | .982 | .982 | **.983** | **.983** | .982 | .980 | .979 | .976 | .975 | .971 | .966 |
| 1 | | .970 | .980 | **.984** | **.984** | **.984** | .983 | .982 | .980 | .978 | .976 | .973 | .968 | .960 |
| 1/128 | Error Rate | **10.44** | 10.56 | 10.96 | 11.86 | 13.50 | 16.16 | 18.26 | 20.50 | 23.44 | 26.95 | 31.39 | 37.92 | 44.54 |
| 1/64 | | 9.67 | **9.29** | 10.23 | 11.04 | 12.29 | 14.55 | 16.52 | 18.58 | 21.40 | 24.78 | 27.65 | 34.12 | 41.67 |
| 1/32 | | 8.87 | **8.66** | 9.44 | 10.35 | 11.29 | 13.59 | 15.34 | 17.30 | 19.31 | 21.82 | 24.86 | 28.37 | 33.91 |
| 1/16 | | 8.19 | **7.92** | 8.93 | 9.67 | 10.37 | 11.93 | 13.51 | 15.35 | 17.42 | 19.40 | 22.30 | 25.74 | 28.36 |
| 1/8 | | 7.59 | **7.32** | 7.87 | 8.65 | 9.26 | 10.31 | 11.63 | 13.06 | 14.68 | 16.39 | 18.28 | 22.50 | 26.87 |
| 1/4 | | 6.87 | **6.44** | 7.04 | 7.49 | 8.01 | 9.05 | 9.86 | 10.56 | 11.45 | 12.28 | 14.36 | 18.05 | 22.59 |
| 1/2 | | 6.04 | **5.71** | 5.97 | 6.45 | 6.66 | 7.14 | 7.53 | 8.03 | 8.80 | 9.94 | 11.44 | 14.85 | 18.37 |
| 3/4 | | 5.81 | **5.31** | 5.48 | 5.75 | 5.87 | 6.25 | 6.57 | 6.89 | 7.58 | 8.72 | 10.69 | 13.92 | 16.29 |
| 1 | | 5.54 | 5.04 | **5.00** | 5.03 | 5.26 | 5.64 | 5.95 | 6.46 | 7.23 | 8.50 | 10.18 | 13.03 | 16.27 |

Table B1a: The Effect of Training-Set Size and Class Distribution on Classifier Performance

Table B1b: The Effect of Training-Set Size and Class Distribution on Classifier Performance

| Data Set | Size | Metric | 2 | 5 | 10 | 20 | 30 | 40 | 47.8* | 50 | 60 | 70 | 80 | 90 | 95 |
|---|---|---|---|---|---|---|---|---|---|---|---|---|---|---|---|
| KR-VS-KP | 1/32 | AUC | .567 | .637 | .680 | .742 | .803 | .852 | .894 | .894 | **.897** | .854 | .797 | .695 | .637 |
| | 1/16 | | .618 | .681 | .800 | .888 | .920 | .942 | .951 | **.952** | .951 | .945 | .929 | .839 | .724 |
| | 1/8 | | .647 | .809 | .893 | .947 | .960 | **.976** | **.976** | **.976** | .975 | .974 | .967 | .936 | .807 |
| | 1/4 | | .768 | .888 | .938 | .980 | .984 | .987 | **.989** | **.989** | **.989** | .985 | .982 | .973 | .947 |
| | 1/2 | | .886 | .946 | .981 | .992 | .994 | **.995** | **.995** | **.995** | **.995** | .994 | .990 | .982 | .974 |
| | 3/4 | | .922 | .966 | .987 | .994 | .995 | **.996** | .995 | **.996** | **.996** | .995 | .994 | .986 | .980 |
| | 1 | | .937 | .970 | .991 | .994 | .997 | **.998** | .997 | **.998** | **.998** | .997 | .994 | .988 | .982 |
| | 1/32 | Error Rate | 42.61 | 36.35 | 33.49 | 27.44 | 21.92 | 17.82 | 14.08 | **14.06** | 17.17 | 21.18 | 26.31 | 33.10 | 38.82 |
| | 1/16 | | 37.99 | 33.02 | 22.76 | 15.49 | 12.66 | 10.46 | 10.14 | **9.74** | 10.08 | 11.53 | 13.97 | 22.14 | 30.95 |
| | 1/8 | | 35.16 | 22.73 | 15.30 | 10.51 | 8.66 | 7.10 | **6.45** | 6.63 | 6.91 | 7.44 | 9.24 | 13.21 | 23.97 |
| | 1/4 | | 26.33 | 15.74 | 11.26 | 6.16 | 5.46 | 4.59 | 4.24 | 4.32 | **4.23** | 5.27 | 5.97 | 8.54 | 12.45 |
| | 1/2 | | 17.11 | 11.07 | 6.00 | 3.71 | 2.72 | 2.38 | 2.05 | **2.11** | 2.32 | 2.66 | 4.16 | 5.61 | 8.66 |
| | 3/4 | | 13.37 | 7.49 | 4.10 | 2.75 | 2.12 | 1.60 | 1.64 | **1.55** | **1.55** | 1.93 | 2.88 | 5.05 | 7.03 |
| | 1 | | 12.18 | 6.50 | 3.20 | 2.33 | 1.73 | **1.16** | 1.39 | 1.22 | 1.34 | 1.53 | 2.55 | 3.66 | 6.04 |

| Data Set | Size | Metric | 2 | 5 | 10 | 20 | 30 | 40 | 40.1* | 50 | 60 | 70 | 80 | 90 | 95 |
|---|---|---|---|---|---|---|---|---|---|---|---|---|---|---|---|
| WEATHER | 1/32 | AUC | .535 | .535 | .535 | .557 | .559 | **.571** | .570 | .570 | .563 | .536 | .556 | .529 | .529 |
| | 1/16 | | .535 | .533 | .562 | .588 | .593 | .595 | .595 | .600 | **.617** | .603 | .597 | .562 | .540 |
| | 1/8 | | .535 | .565 | .591 | .617 | .632 | **.651** | **.651** | .642 | .619 | .617 | .615 | .583 | .555 |
| | 1/4 | | .563 | .606 | .627 | .680 | .678 | .670 | .670 | .671 | .672 | **.675** | .644 | .615 | .600 |
| | 1/2 | | .578 | .626 | .682 | .690 | .705 | **.712** | **.712** | .707 | .700 | .690 | .679 | .664 | .629 |
| | 3/4 | | .582 | .657 | .698 | .700 | .715 | **.720** | **.720** | .713 | .711 | .699 | .700 | .661 | .642 |
| | 1 | | .694 | .715 | .728 | .737 | .738 | **.740** | .736 | .736 | .730 | .736 | .722 | .718 | .702 |
| | 1/32 | Error Rate | **40.76** | **40.76** | **40.76** | 41.06 | 41.55 | 41.91 | 41.91 | 45.40 | 49.73 | 48.59 | 52.77 | 53.77 | |
| | 1/16 | | 39.56 | 39.56 | **38.85** | 38.87 | 39.96 | 39.81 | 39.81 | 41.19 | 41.08 | 43.25 | 46.42 | 52.73 | 53.55 |
| | 1/8 | | 39.27 | 38.70 | 37.95 | **36.45** | 37.01 | 37.68 | 39.29 | 41.49 | 42.84 | 46.32 | 51.34 | 53.46 | |
| | 1/4 | | 39.00 | 38.11 | 36.72 | **35.40** | 35.84 | 36.98 | 36.98 | 37.79 | 38.37 | 40.47 | 45.47 | 50.68 | 53.32 |
| | 1/2 | | 38.62 | 37.66 | 35.89 | 35.32 | **34.39** | 35.62 | 35.62 | 36.47 | 37.62 | 40.07 | 44.11 | 49.80 | 53.19 |
| | 3/4 | | 38.56 | 37.23 | 35.38 | 35.23 | **34.14** | 34.25 | 35.21 | 36.08 | 37.35 | 39.91 | 43.55 | 49.46 | 52.53 |
| | 1 | | 38.41 | 36.89 | 35.25 | 33.68 | **33.11** | 33.43 | 33.69 | 34.61 | 36.09 | 38.36 | 41.68 | 47.23 | 51.69 |

| Data Set | Size | Metric | 2 | 3.9* | 5 | 10 | 20 | 30 | 40 | 50 | 60 | 70 | 80 | 90 | 95 |
|---|---|---|---|---|---|---|---|---|---|---|---|---|---|---|---|
| LETTER-A | 1/32 | AUC | .532 | .532 | .532 | .558 | .637 | .699 | .724 | **.775** | .765 | .769 | .745 | .747 | .724 |
| | 1/16 | | .552 | .601 | .601 | .639 | .704 | .726 | .798 | .804 | .828 | **.833** | .830 | .799 | .780 |
| | 1/8 | | .603 | .622 | .642 | .645 | .758 | .798 | .826 | .841 | .860 | .861 | **.871** | .854 | .824 |
| | 1/4 | | .637 | .654 | .692 | .743 | .793 | .845 | .865 | .878 | .893 | .899 | **.904** | .900 | .876 |
| | 1/2 | | .677 | .724 | .734 | .790 | .868 | .893 | .912 | .916 | .921 | .926 | **.933** | .927 | .910 |
| | 3/4 | | .702 | .745 | .776 | .841 | .890 | .908 | .917 | .930 | .935 | .941 | **.948** | .939 | .927 |
| | 1 | | .711 | .772 | .799 | .865 | .891 | .911 | .938 | .937 | .944 | .951 | **.954** | .952 | .940 |
| | 1/32 | Error Rate | **7.86** | **7.86** | **7.86** | 8.81 | 11.11 | 12.58 | 12.31 | 15.72 | 19.66 | 22.55 | 32.06 | 42.38 | 48.52 |
| | 1/16 | | **5.19** | 6.04 | 6.04 | 7.38 | 8.05 | 9.23 | 10.48 | 14.44 | 16.40 | 20.84 | 27.38 | 40.64 | 47.61 |
| | 1/8 | | 4.60 | **4.58** | 4.84 | 5.22 | 6.76 | 8.19 | 10.03 | 12.32 | 13.67 | 16.74 | 24.00 | 35.44 | 45.09 |
| | 1/4 | | 4.38 | **4.36** | 4.77 | 5.25 | 6.12 | 6.87 | 7.90 | 9.66 | 12.21 | 14.33 | 18.69 | 30.22 | 43.12 |
| | 1/2 | | 3.63 | 3.49 | **3.47** | 3.97 | 4.27 | 5.32 | 6.08 | 7.03 | 9.02 | 10.33 | 15.65 | 22.76 | 35.93 |
| | 3/4 | | 3.22 | 3.08 | 3.07 | **3.05** | 3.36 | 4.04 | 5.23 | 5.99 | 7.31 | 9.86 | 12.93 | 20.60 | 29.62 |
| | 1 | | 2.86 | 2.78 | 2.75 | **2.59** | 3.03 | 3.79 | 4.53 | 5.38 | 6.48 | 8.51 | 12.37 | 18.10 | 26.14 |

| Data Set | Size | Metric | 2 | 5 | 10 | 20 | 30 | 35.6* | 40 | 50 | 60 | 70 | 80 | 90 | 95 |
|---|---|---|---|---|---|---|---|---|---|---|---|---|---|---|---|
| BLACKJACK | 1/32 | AUC | .545 | .575 | .593 | .607 | .620 | .621 | **.624** | .619 | .618 | .609 | .600 | .580 | .532 |
| | 1/16 | | .556 | .589 | .603 | .613 | .629 | .636 | .643 | **.651** | .648 | .634 | .622 | .594 | .551 |
| | 1/8 | | .579 | .592 | .604 | .639 | .651 | .657 | .657 | **.665** | **.665** | .659 | .630 | .603 | .553 |
| | 1/4 | | .584 | .594 | .612 | .652 | .672 | .673 | .677 | **.686** | **.686** | .680 | .650 | .603 | .554 |
| | 1/2 | | .587 | .596 | .621 | .675 | .688 | .692 | .697 | .703 | **.704** | .690 | .670 | .603 | .556 |
| | 3/4 | | .593 | .596 | .622 | .675 | .688 | .699 | .703 | **.710** | **.710** | .699 | .677 | .604 | .558 |
| | 1 | | .593 | .596 | .628 | .678 | .688 | .700 | .712 | .713 | **.715** | .700 | .678 | .604 | .558 |
| | 1/32 | Error Rate | 34.26 | 33.48 | 32.43 | 32.30 | **31.97** | 32.44 | 32.84 | 33.48 | 34.89 | 36.05 | 38.04 | 38.31 | 43.65 |
| | 1/16 | | 34.09 | 32.96 | 31.27 | **30.41** | 30.57 | 30.91 | 30.97 | 31.82 | 32.12 | 33.61 | 35.55 | 38.19 | 37.86 |
| | 1/8 | | 32.83 | 31.90 | 30.70 | **29.63** | 29.71 | 30.02 | 30.30 | 30.66 | 31.34 | 32.05 | 32.44 | 35.11 | 37.73 |
| | 1/4 | | 31.84 | 30.78 | 30.60 | 29.61 | **29.25** | 29.34 | 29.64 | 29.62 | 30.40 | 30.86 | 31.33 | 33.02 | 35.09 |
| | 1/2 | | 31.11 | 30.70 | 30.30 | 28.96 | 28.73 | **28.60** | 29.03 | 29.33 | 29.32 | 30.10 | 31.32 | 32.80 | 34.46 |
| | 3/4 | | 30.80 | 30.68 | 29.93 | 28.73 | 28.56 | **28.44** | 28.50 | 28.77 | 28.99 | 29.95 | 31.17 | 32.75 | 34.18 |
| | 1 | | 30.74 | 30.66 | 29.81 | 28.67 | 28.56 | **28.40** | 28.45 | 28.71 | 28.91 | 29.78 | 31.02 | 32.67 | 33.87 |

Table B1b: The Effect of Training-Set Size and Class Distribution on Classifier Performance

The benefit of selecting the class distribution of the training data is demonstrated using several examples. Table B1a highlights six cases (by using a line to connect pairs of data points) where competitive or improved performance is achieved from fewer training examples. In each of these six cases, the data point corresponding to the smaller data-set size performs as well or better than the data point that corresponds to the larger data-set size (the latter being either the natural distribution or a balanced one).

## Appendix C: Detailed Results for the Budget-Sensitive Sampling Algorithm

This appendix describes the execution of the progressive sampling algorithm that was described in Table 7. The execution of the algorithm is evaluated using the detailed results from Appendix B. First, in Table C1, a detailed iteration-by-iteration description of the sampling algorithm is presented as it is applied to the phone data set using error rate to measure classifier performance. Table C2 then provides a more compact version of this description, by reporting only the key variables as they change value from iteration to iteration. Finally, in Table C3a and Table C3b, this compact description is used to describe the execution of the sampling algorithm for the phone, adult, covertype, kr-vs-kp, weather and blackjack data sets, using both error rate and AUC to measure performance. Note that for each of these tables, the column labeled "budget" refers to the budget used, or cost incurred—and that in no case is the budget exceeded, which means that all examples requested during the execution of the algorithm are used in the final training set, with the heuristically-determined class distribution (i.e., the algorithm is budget-efficient).

The results that are described in this appendix, consistent with the results presented in Section 7, are based on a geometric factor, $\mu$, of 2, and a value of $cmin$ of 1/32. The total budget available for procuring training examples is $n$. Based on these values, the value of $K$, which determines the number of iterations of the algorithm and is computed on line 2 of Table 7, is set to 5. Note that the value of $n$ is different for each data set and, given the methodology for altering the class distribution specified in Section 4.1, if the training set size in Table 2 is S and the fraction of minority-class examples is $f$, then $n = \frac{3}{4} \cdot S \cdot f$.

Below is the description of the sampling algorithm, as it is applied to the phone data set with error rate as the performance measure:

---

$\underline{j = 0}$  Training-set size = 1/32 $n$. Form 13 data sets, which will contain between 2% and 95% minority-class examples. This requires .0297$n$ (95% of 1/32 $n$) minority-class examples and .0306$n$ (100%-2% = 98% of 1/32 $n$) majority-class examples. Induce and then evaluate the resulting classifiers. Based on the results in Table 7, the natural distribution, which contains 18.2% minority-class examples, performs best. Total Budget: .0603$n$ (.0297$n$ minority, .0306$n$ majority).

$\underline{j = 1}$  Training-set size = 1/16 $n$. Form data sets corresponding to the best-performing class distribution form the previous iteration (18.2% minority) and the adjoining class distributions used in the beam search, which contain 10% and 20% minority-class examples. This requires .0250$n$ (20% of 1/16 $n$) minority-class examples and .0563$n$ (90% of 1/16 $n$) majority-class examples. Since .0297$n$ minority-class examples were previously obtained, class distributions containing 30% and 40% minority-class examples can also be formed without requesting additional examples. This iteration requires $.0257n$ additional majority-class examples. The best-performing distribution contains 10% minority-class examples. Total Budget: .0860 $n$ (.0297$n$ minority, .0563$n$ majority).

<u>j = 2</u>   Training-set size = 1/8 $n$. Since the 10% distribution performed best, the beam search evaluates the 5%, 10%, and 18.2% minority-class distributions. The 20% class distribution is also evaluated since this requires only $.0250n$ of the $.0297n$ previously obtained minority-class examples. A total of $.1188n$ (95% of 1/8 $n$) majority-class examples are required. The best performing distribution contains 10% minority-class examples. This iteration requires $.0625n$ additional majority-class examples. Total Budget: .1485$n$ (.0297$n$ minority, .1188$n$ majority).

<u>j = 3</u>   Training-set size = 1/4 $n$. The distributions to be evaluated are 5%, 10%, and 18.2%. There are no "extra" minority-class examples available to evaluate additional class distributions. This iteration requires $.0455n$ (18.2% of 1/4 $n$) minority-class examples and $.2375n$ (95% of 1/4 $n$) majority-class examples. The best-performing class distribution contains 10% minority-class examples. Total Budget: .2830$n$ (.0455$n$ minority, .2375$n$ majority)

<u>j = 4</u>   Training-set size = 1/2 $n$. The 5%, 10%, and 18.2% class distributions are evaluated. This iteration requires $.0910n$ (18.2% of 1/2 $n$) minority-class examples and $.4750n$ (95% of 1/2 $n$) majority-class examples. The best-performing distribution contains 10% minority-class examples. Total Budget: .5660$n$ (.0910$n$ minority, .4750$n$ majority).

<u>j = 5</u>   Training-set size = $n$. For this last iteration only the best class distribution from the previous iteration is evaluated. Thus, a data set of size $n$ is formed, containing $.1n$ minority-class examples and $.9n$ majority-class examples. Thus $.0090n$ additional minority-class examples and $.4250n$ additional majority-class examples are required. Since all the previously obtained examples are used, there is no "waste" and the budget is not exceeded. Total Budget: 1.0$n$ (.1000$n$ minority, .9000$n$ majority)

Table C1: A Detailed Example of the Sampling Algorithm (Phone Data Set using Error Rate)

| | | | | | Expressed as a fraction on $n$ | | | |
|---|---|---|---|---|---|---|---|---|
| j | size | class-distr | best | min-need | maj-need | minority | majority | budget |
| 0 | 1/32 $n$ | all | 18.2% | .0297 | .0306 | .0297 | .0306 | .0603 |
| 1 | 1/16 $n$ | 10, *18.2*, 20, 30, 40 | 10% | .0250 | .0563 | .0297 | .0563 | .0860 |
| 2 | 1/8 $n$ | 5, *10*, 18.2, 20 | 10% | .0250 | .1188 | .0297 | .1188 | .1485 |
| 3 | 1/4 $n$ | 5, *10*, 18.2 | 10% | .0455 | .2375 | .0455 | .2375 | .2830 |
| 4 | 1/2 $n$ | 5, *10*, 18.2 | **10%** | .0910 | .4750 | .0910 | .4750 | .5660 |
| 5 | 1 $n$ | 10 | | .1000 | .9000 | .1000 | .9000 | 1.0000 |

Table C2: Compact Description of the Results in Table B1a for the Phone Data Set

| Data set | Metric | j | size | class-distr | best | Expressed as a fraction of *n* | | | | |
|---|---|---|---|---|---|---|---|---|---|---|
| | | | | | | min-need | maj-need | minority | majority | budget |
| Phone | ER | 0 | 1/32 n | all | 18.2 | .0297 | .0306 | .0297 | .0306 | .0603 |
| | | 1 | 1/16 n | 10, *18.2*, 20, 30, 40 | 10 | .0250 | .0563 | .0297 | .0563 | .0860 |
| | | 2 | 1/8 n | 5, *10*, 18.2, 20 | 10 | .0250 | .1188 | .0297 | .1188 | .1485 |
| | | 3 | 1/4 n | 5, *10*, 18.2 | 10 | .0455 | .2375 | .0455 | .2375 | .2830 |
| | | 4 | 1/2 n | 5, *10*, 18.2 | **10** | .0910 | .4750 | .0910 | .4750 | .5660 |
| | | 5 | 1 n | 10 | | .1000 | .9000 | .1000 | .9000 | 1.0 |
| Phone | AUC | 0 | 1/32 n | all | 20 | .0297 | .0306 | .0297 | .0306 | .0603 |
| | | 1 | 1/16 n | 18.2, *20*, 30, 40 | 30 | .0250 | .0511 | .0297 | .0511 | .0808 |
| | | 2 | 1/8 n | 20, *30*, 40 | 30 | .0500 | .1000 | .0500 | .1000 | .1500 |
| | | 3 | 1/4 n | 20, *30*, 40 | 20 | .1000 | .2000 | .1000 | .2000 | .3000 |
| | | 4 | 1/2 n | 18.2, *20*, 30 | **18.2** | .1500 | .4090 | .1500 | .4090 | .5590 |
| | | 5 | 1 n | 18.2 | | .1820 | .8180 | .1820 | .8180 | 1.0 |
| Adult | ER | 0 | 1/32 n | all | 20 | .0297 | .0306 | .0297 | .0306 | .0603 |
| | | 1 | 1/16 n | 10, *20*, 23.9, 30, 40 | 20 | .0250 | .0563 | .0297 | .0563 | .0860 |
| | | 2 | 1/8 n | 10, *20*, 23.9 | 20 | .0299 | .1125 | .0299 | .1125 | .1424 |
| | | 3 | 1/4 n | 10, *20*, 23.9 | 10 | .0598 | .2250 | .0598 | .2250 | .2848 |
| | | 4 | 1/2 n | 5, *10*, 20 | **20** | .1000 | .4750 | .1000 | .4750 | .5750 |
| | | 5 | 1 n | 20 | | .2000 | .8000 | .2000 | .8000 | 1.0 |
| Adult | AUC | 0 | 1/32 n | all | 80 | .0297 | .0306 | .0297 | .0306 | .0603 |
| | | 1 | 1/16 n | 60, 70, *80*, 90 | 70 | .0563 | .0250 | .0563 | .0306 | .0869 |
| | | 2 | 1/8 n | 60, *70*, 80 | 70 | .1000 | .0500 | .1000 | .0500 | .1500 |
| | | 3 | 1/4 n | 60, *70*, 80 | 80 | .2000 | .1000 | .2000 | .1000 | .3000 |
| | | 4 | 1/2 n | 70, *80*, 90 | **80** | .4500 | .1500 | .4500 | .1500 | .6000 |
| | | 5 | 1 n | 80 | | .8000 | .2000 | .8000 | .2000 | 1.0 |
| Covertype | ER | 0 | 1/32 n | all | 5 | .0297 | .0306 | .0297 | .0306 | .0603 |
| | | 1 | 1/16 n | 2, *5*, 10, 20, 30, 40 | 5 | .0250 | .0613 | .0297 | .0613 | .0910 |
| | | 2 | 1/8 n | 2, *5*, 10, 20 | 5 | .0250 | .1225 | .0297 | .1225 | .1522 |
| | | 3 | 1/4 n | 2, *5*, 10 | 5 | .0250 | .2450 | .0297 | .2450 | .2747 |
| | | 4 | 1/2 n | 2, *5*, 10 | **5** | .0500 | .4900 | .0500 | .4900 | .5400 |
| | | 5 | 1 n | 5 | | .0500 | .9500 | .0500 | .9500 | 1.0 |
| Covertype | AUC | 0 | 1/32 n | all | 20 | .0297 | .0306 | .0297 | .0306 | .0603 |
| | | 1 | 1/16 n | 14.8, *20*, 30, 40 | 30 | .0250 | .0533 | .0297 | .0533 | .0830 |
| | | 2 | 1/8 n | 20, *30*, 40 | 40 | .0500 | .1000 | .0500 | .1000 | .1500 |
| | | 3 | 1/4 n | 30, *40*, 50 | 30 | .1250 | .1750 | .1250 | .1750 | .3000 |
| | | 4 | 1/2 n | 20, *30*, 40 | **20** | .2000 | .4000 | .2000 | .4000 | .6000 |
| | | 5 | 1 n | 20 | | .2000 | .8000 | .2000 | .8000 | 1.0 |
| Kr-vs-kp | ER | 0 | 1/32 *n* | all | 50 | .0297 | .0306 | .0297 | .0306 | .0603 |
| | | 1 | 1/16 *n* | 47.8, *50*, 60 | 50 | .0375 | .0327 | .0375 | .0327 | .0702 |
| | | 2 | 1/8 *n* | 47.8, *50*, 60 | 47.8 | .0750 | .0653 | .0750 | .0653 | .1403 |
| | | 3 | 1/4 *n* | 40, *47.8*, 50 | 47.8 | .1250 | .1500 | .1250 | .1500 | .2750 |
| | | 4 | 1/2 *n* | 40, *47.8*, 50 | **50** | .2500 | .3000 | .2500 | .3000 | .5500 |
| | | 5 | 1 *n* | 50 | | .5000 | .5000 | .5000 | .5000 | 1.0 |
| Kr-vs-kp | AUC | 0 | 1/32 *n* | all | 60 | .0297 | .0306 | .0297 | .0306 | .0603 |
| | | 1 | 1/16 *n* | 50, *60*, 70 | 50 | .0438 | .0313 | .0438 | .0313 | .0751 |
| | | 2 | 1/8 *n* | 47.8, *50*, 60 | 50 | .0750 | .0653 | .0750 | .0653 | .1403 |
| | | 3 | 1/4 *n* | 47.8, *50*, 60 | 50 | .1500 | .1305 | .1500 | .1305 | .2805 |
| | | 4 | 1/2 *n* | 47.8, *50*, 60 | **50** | .3000 | .2610 | .3000 | .2610 | .5610 |
| | | 5 | 1 *n* | 50 | | .5000 | .5000 | .5000 | .5000 | 1.0 |

Table C3a: Summary Results for the Sampling Algorithm (phone, adult, covertype, kr-vs-kp)

| Data set | Metric | j | size | class-distr | best | Expressed as a fraction of n | | | | |
|---|---|---|---|---|---|---|---|---|---|---|
| | | | | | | min-need | maj-need | minority | majority | budget |
| Weather | ER | 0 | 1/32 n | all | 5 | .0297 | .0306 | .0297 | .0316 | .0613 |
| | | 1 | 1/16 n | 2,5,10,20,30,40, 40.1 | 10 | .0250 | .0613 | .0297 | .0613 | .0910 |
| | | 2 | 1/8 n | 5, 10, 20 | 20 | .0250 | .1188 | .0297 | .1188 | .1485 |
| | | 3 | 1/4 n | 10, 20, 30 | 20 | .0750 | .2250 | .0750 | .2250 | .3000 |
| | | 4 | 1/2 n | 10, 20, 30 | **30** | .1500 | .4500 | .1500 | .4500 | .6000 |
| | | 5 | 1 n | 30 | | .3000 | .7000 | .3000 | .7000 | 1.0 |
| Weather | AUC | 0 | 1/32 n | all | 40 | .0297 | .0306 | .0297 | .0316 | .0613 |
| | | 1 | 1/16 n | 30, 40, 50 | 50 | .0313 | .0438 | .0313 | .0438 | .0751 |
| | | 2 | 1/8 n | 40, 50, 60 | 40 | .0750 | .0750 | .0750 | .0750 | .1500 |
| | | 3 | 1/4 n | 30, 40, 50 | 50 | .1250 | .1750 | .1250 | .1750 | .3000 |
| | | 4 | 1/2 n | 40, 50, 60 | **40** | .3000 | .3000 | .3000 | .3000 | .6000 |
| | | 5 | 1 n | 40 | | .4000 | .6000 | .4000 | .6000 | 1.0 |
| Letter-a | ER | 0 | 1/32 n | all | 3.9 | .0297 | .0306 | .0297 | .0306 | .0603 |
| | | 1 | 1/16 n | 2, 3.9, 5, 10, 20, 30, 40 | 2 | .0250 | .0613 | .0297 | .0613 | .0910 |
| | | 2 | 1/8 n | 2, 3.9, 5, 10, 20 | 3.9 | .0250 | .1125 | .0297 | .1225 | .1522 |
| | | 3 | 1/4 n | 2, 3.9, 5, 10 | 3.9 | .0250 | .2450 | .0297 | .2450 | .2747 |
| | | 4 | 1/2 n | 2, 3.9, 5 | **5** | .0250 | .4900 | .0250 | .4900 | .5150 |
| | | 5 | 1 n | 5 | | .0500 | .9500 | .0500 | .9500 | 1.0 |
| Letter-a | AUC | 0 | 1/32 n | all | 50 | .0297 | .0306 | .0297 | .0306 | .0603 |
| | | 1 | 1/16 n | 40, 50, 60 | 60 | .0375 | .0375 | .0375 | .0375 | .0750 |
| | | 2 | 1/8 n | 50, 60, 70 | 70 | .0875 | .0625 | .0875 | .0625 | .1500 |
| | | 3 | 1/4 n | 60, 70, 80 | 80 | .2000 | .1000 | .2000 | .1000 | .3000 |
| | | 4 | 1/2 n | 70, 80, 90 | **80** | .4500 | .1500 | .4500 | .1500 | .6000 |
| | | 5 | 1 n | | | .8000 | .2000 | .8000 | .2000 | 1.0 |
| Blackjack | ER | 0 | 1/32 n | all | 30 | .0297 | .0306 | .0297 | .0316 | .0613 |
| | | 1 | 1/16 n | 20, 30, 35.6, 40 | 20 | .0250 | .0500 | .0297 | .0500 | .0797 |
| | | 2 | 1/8 n | 10, 20, 30 | 20 | .0375 | .1125 | .0375 | .1125 | .1500 |
| | | 3 | 1/4 n | 10, 20, 30 | 30 | .0750 | .2250 | .0750 | .2250 | .3000 |
| | | 4 | 1/2 n | 20, 30, 35.6 | **35.6** | .1780 | .4000 | .1780 | .4000 | .5780 |
| | | 5 | 1 n | 35.6 | | .3560 | .6440 | .3560 | .6440 | 1.0 |
| Blackjack | AUC | 0 | 1/32 n | all | 40 | .0297 | .0306 | .0297 | .0316 | .0613 |
| | | 1 | 1/16 n | 20, 35.6, 40, 50 | 50 | .0313 | .0403 | .0313 | .0403 | .0716 |
| | | 2 | 1/8 n | 40, 50, 60 | 50 | .0750 | .0750 | .0750 | .0750 | .1500 |
| | | 3 | 1/4 n | 40, 50, 60 | 50 | .1500 | .1500 | .1500 | .1500 | .3000 |
| | | 4 | 1/2 n | 40, 50, 60 | **60** | .3000 | .3000 | .3000 | .3000 | .6000 |
| | | 5 | 1 n | 60 | | .6000 | .4000 | .6000 | .4000 | 1.0 |

Table C3b: Summary Results for the Sampling Algorithm (weather, letter-a, blackjack)